\documentclass[a4paper,fleqn]{cas-dc}

\usepackage{soul}
\soulregister{\cite}7
\soulregister{\ref}7

\usepackage[numbers]{natbib}

\def\tsc#1{\csdef{#1}{\textsc{\lowercase{#1}}\xspace}}
\tsc{WGM}
\tsc{QE}

\begin{document}
\let\WriteBookmarks\relax
\def\floatpagepagefraction{1}
\def\textpagefraction{.001}
\let\printorcid\relax 

\shorttitle{Middle Fusion and Multi-Stage, Multi-Form Prompts for Robust RGB-T Tracking}    

\shortauthors{Q. Wang et al.}

\title[mode = title]{Middle Fusion and Multi-Stage, Multi-Form Prompts for Robust RGB-T Tracking}

\author[1]{Qiming Wang}[]
\credit{Conceptualization, Investigation, Methodology, Software, Visualization, Validation, Writing – original draft.}

\author[1]{Yongqiang Bai}
\cormark[1] 
\ead{byfengyun@bit.edu.cn} 
\credit{Funding acquisition, Supervision, Project administration, Writing – review & editing.}

\author[2]{Hongxing Song}
\credit{Resources, Data curation.}

\address[1]{National Key Lab of Autonomous Intelligent Unmanned Systems, Beijing Institute of Technology, Beijing 100081, China}
\address[2]{North Lianchuang Communication Company, Jiangxi 600363, China}

\cortext[1]{Corresponding author} 

\begin{abstract}
RGB-T tracking, a vital downstream task of object tracking, has made remarkable progress in recent years. Yet, it remains hindered by two major challenges: 1) the trade-off between performance and efficiency; 2) the scarcity of training data. To address the latter challenge, some recent methods employ prompts to fine-tune pre-trained RGB tracking models and leverage upstream knowledge in a parameter-efficient manner. However, these methods inadequately explore modality-independent patterns and disregard the dynamic reliability of different modalities in open scenarios. We propose M3PT, a novel RGB-T prompt tracking method that leverages middle fusion and multi-modal and multi-stage visual prompts to overcome these challenges. We pioneer the use of the adjustable middle fusion meta-framework for RGB-T tracking, which could help the tracker balance the performance with efficiency, to meet various demands of application. Furthermore, based on the meta-framework, we utilize multiple flexible prompt strategies to adapt the pre-trained model to comprehensive exploration of uni-modal patterns and improved modeling of fusion-modal features in diverse modality-priority scenarios, harnessing the potential of prompt learning in RGB-T tracking. Evaluating on 6 existing challenging benchmarks, our method surpasses previous state-of-the-art prompt fine-tuning methods while maintaining great competitiveness against excellent full-parameter fine-tuning methods, with only 0.34M fine-tuned parameters.
\end{abstract}




\maketitle

\section{Introduction}

Visual object tracking is a fundamental task in the computer vision community, which aims to continuously predict the state of a given target in every frame of a video sequence. It has a wide range of applications in fields such as autonomous driving \cite{autonomous}, robot perception \cite{robot}, and intelligent surveillance \cite{surveillance}. However, the tracking methods based on visible modality (also known as RGB tracking) are not robust in challenging scenarios such as extreme illumination and adverse weather, where the visible image quality is poor. To solve this problem, researchers have developed RGB-T tracking methods \cite{2023survey, 2022survey}, which use the complementary information between thermal infrared and visible modalities to achieve all-weather and all-day tracking.

\begin{figure}
    \centering
    \includegraphics[width=1\linewidth]{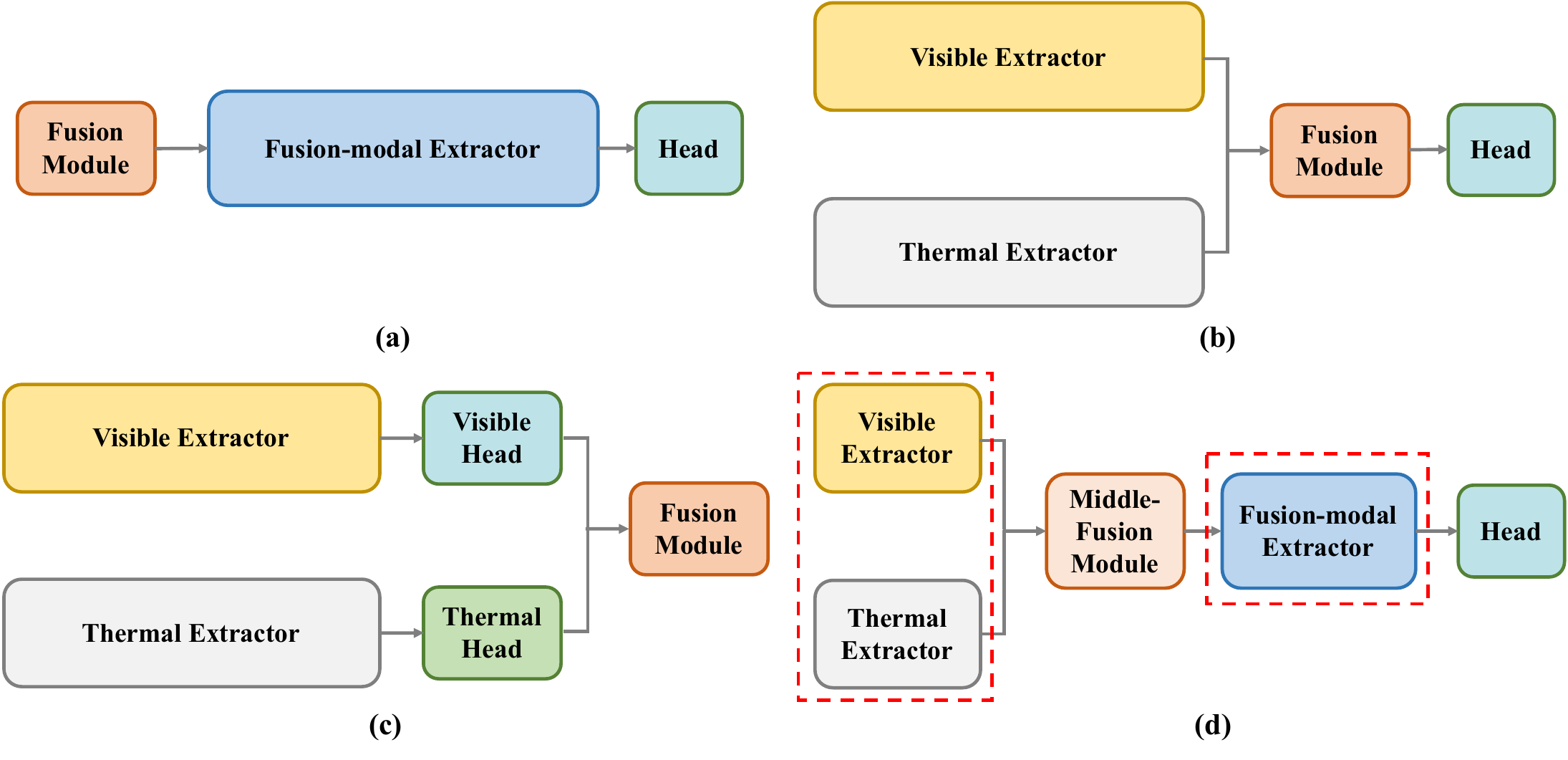}
    \caption{Comparison of different fusion tracking frameworks. (a)-(c) are the three mainstream frameworks: image-level, feature-level, and decision-level multi-modal fusion tracking frameworks, respectively. (d) is our proposed middle fusion meta-framework. Our framework, unlike the other three, splits the backbone into dual-stream and single-stream structures for uni-modal and fusion-modal feature representation, respectively, with the fusion module situated between them.}
    \label{framework}
\end{figure}

RGB-T tracking data has a complex and rich cross-modal feature space, which is a correlated combination of the two modal feature spaces rather than a simple addition. Since the data of the two modalities are synchronized in time and space, they share multiple patterns such as object boundaries, partial fine-grained information, etc.; however, due to the different imaging bands, they also have many modality-specific and discriminative patterns. These patterns form the key information in the feature space, and exploring, integrating, and utilizing them is essential for robust RGB-T tracking. To achieve this goal, most excellent works \cite{manet, mfdimp, cat, m5l, manet++, jmmac, siamcda, tfnet, adrnet, apfnet, mfnet, vtuav, drgcnet, siamafts, siammlaa, siamtdr, tbsi, cmd} are built on various effective fusion-based multi-modal tracking meta-frameworks. These meta-frameworks can be classified into three categories based on the fusion location: image-level, feature-level, and decision-level fusion frameworks, as shown in Figure \ref{framework}(a)-(c). The image-level fusion framework is simple and efficient, but it does not exploit the two modalities separately, leading to poor performance. On the other hand, the feature-level and decision-level fusion frameworks extract rich features from both modalities, achieving high performance, but they introduce redundant computations due to the dual-stream architecture, which greatly reduces the inference efficiency. Therefore, a natural question arises: can we design a general RGB-T fusion tracking framework that enables trackers utilizing this framework to better balance efficiency with performance?

Inspired by this idea, we develop a general middle fusion meta-framework for RGB-T tracking, as shown in Figure \ref{framework}(d). In this meta-framework, we place the fusion module in the middle of the backbone, and split the backbone into two parts: the pre-fusion backbone extracts the information of the two modalities independently, and the post-fusion backbone enhances the fusion-modal features. By choosing an appropriate fusion location and utilize effective fusion module, the trackers built on this mixed-stream meta-framework can achieve effective uni-modal modeling, as well as reduce the redundant feature extraction process, thus achieving a good balance between performance and efficiency. Also, by adjusting the fusion location, the efficiency and performance of the trackers could be changed flexibly for diverse demands. To the best of our knowledge, this is the first time that the middle fusion meta-framework has been introduced into the RGB-T tracking task.

Another major focus of current research is how to address the data scarcity problem. The high cost of data acquisition leads to a persistent lack of sufficient annotated training data for the RGB-T tracking task, which severely hinders the data-driven model from learning reliable task-specific knowledge. Since most RGB-T tracking methods are derived from upstream methods, some studies \cite{ mfdimp, vtuav, rgbt234, cmd, siamivfn, tbsi, ustrack} use pre-trained upstream models to initialize the parameters and facilitate faster and better convergence of the current model. However, these full fine-tuning methods require tuning a large number of parameters, which imposes extremely high demands on the computing and storage capabilities of the device, and may cause overfitting and forgetting of upstream knowledge.

Considering these issues, some recent works \cite{protrack, vipt, untrack, onetracker} introduce prompt learning into RGB-T tracking to better leverage the inheritance between upstream and downstream tasks. By tuning a minimal number of parameters, these methods effectively narrow the gap between upstream and downstream tasks, delivering competitive performance akin to most full fine-tuning approaches, while significantly reducing training costs. However, existing prompt-based methods have several key limitations: (i) Through direct or indirect image-level fusion, these methods hardly exploit the modality-independent information, resulting in insufficient exploration and utilization of the discriminative patterns within the modality. (ii) They treat thermal infrared modality as an extra information only for generating prompts and simply sum two modality features through constant global or element-wise weights, which neglect the dynamic modality-priority in intricate scenarios, thus failing to fully leverage the advantages of RGB-T tracking. (iii) They generate and inject prompts in a single manner like element-wise summation, which limits the great potential of prompt learning.

Motivated by above discussions, we aim to employ adaptable visual prompts to fine-tune the pre-trained upstream model to identify and leverage the distinctive and supplementary information within multi-modal features, thus optimizing the performance of RGB-T prompt tracking. For this purpose, we introduce four innovative visual prompt strategies to transfer the upstream knowledge, based the middle fusion meta-framework. Consequently, we devise an novel and parameter-efficient RGB-T prompt tracking method with middle fusion and multi-stage, multi-form prompts (M3PT). In the method, we divide the backbone of the foundation model into two parts, each for one of the two modeling stages of the middle fusion framework.
To adapt the first part to effective uni-modal and cross-modal modeling, we propose the Uni-modal and Inter-modal Exploration Prompt Strategy, which uses the first part backbone and two types of lightweight prompters UEP and IP to explore the modality-independent and cross-modal complementary patterns layer by layer, and further operate reliable intra-modal and inter-modal prompting to guide the subsequent layers to model key features of the current modality and different modalities. Considering the dynamic complementarity of different modalities in open scenarios, we design a Middle Fusion Prompt Strategy to achieve adaptive selection and complementary fusion of the discriminative features of the two modalities, and use the fusion-modal features as natural prompts, which are fed into the second-stage backbone.
To better adapt the backbone to the fusion-modal feature modeling, we design a Fusion-modal Enhancement Prompt Strategy, which guides the backbone to extract sufficient global and local fusion features, and thus obtain a richer fusion modality representation. Motivated by VPT, we design a Stage-aware and Modality-aware Prompt Strategy, which generates multiple learnable prompts to store the modality-fixed patterns and prepend them to the input of each stage. This strategy provides clear modality and stage indicators for the backbone. Notably, our method only contains 0.34M fine-tuning parameters. On six challenging and large-scale benchmarks, our method surpasses all previous state-of-the-art prompt fine-tuning methods while maintaining great competitiveness against excellent full-parameter fine-tuning methods, with the least fine-tuned parameters. Our work has the following main contributions:

(1) We propose a flexible and general middle fusion meta-framework for RGB-T tracking task for the first time. Building on this framework, the tracker could easily achieve a good balance between performance and efficiency, to meet diverse application demands.

(2) Built on the proposed middle fusion meta-framework, we develop a novel RGB-T prompt tracking method with multi-form and multi-stage visual prompts (M3PT), which fully unleashes the great potential of RGB-T prompt tracking through four flexible novel visual prompt strategies.

(3) Extensive experiments on six challenging RGB-T tracking benchmarks validate the effectiveness and parameter-efficiency of our method.

\section{Related Work}

\subsection{RGB-T Tracking}

In recent years, many excellent works \cite{siamfc, siamrpn, siamrpn++, dimp, siamcar, transtrack, stark, hift, swintrack, ostrack, mixformer, simtrack, tatrack, romtrack} have been produced on RGB object tracking. However, these methods are limited by the low image quality of visible modality in some challenging scenarios such as illumination changes, extreme lighting, and adverse weather, failing to meet the application requirements of tracking methods in open scenarios. Therefore, researchers have proposed and attracted widespread attention to RGB-T tracking, which utilizes the complementarity of visible and thermal infrared modalities to achieve all-weather tracking.

Lu et al. \cite{manet++} employ diverse adapters to capture the intra-modal and inter-modal relation of different modalities, and propose a new loss function that reduces the distribution divergence of multi-modal features across layers. Zhang et al. \cite{jmmac} develop a late fusion method that achieves robust multi-modal fusion through global and local weights, and uses Kalman filtering for motion estimation, to exploit the joint appearance and motion cues. Zhang et al. \cite{siamcda} extend the structure of siamese-like RGB trackers and introduce a weight generation model to enable the selection of uni-modal discriminative features and the interaction of bimodal discriminative features. Tu et al. \cite{m5l} propose a novel multi-modal multi-margin metric learning framework for RGB-T tracking, which preserves the relations of multilevel hard samples during training. A major challenge for data-driven models in RGB-T tracking is the scarcity of labeled training data, which results from the high cost of data collection, alignment, and annotation. Compared to the abundant training data \cite{got10k, lasot, trackingnet} for RGB tracking task, the training data \cite{vtuav, lasher} for RGB-T tracking is severely inadequate for learning sufficient task-relevant knowledge.

To address the problem of insufficient task data, some researchers propose attribute-based methods \cite{cat, adrnet, apfnet}, which configure lightweight branches for each attribute to model the target appearance under these attributes, and then fuse these appearance features. During the training phase, these methods select sequences of each attribute from the training set to train the corresponding branches, thereby reducing the model’s dependence on the training data to some extent. However, these methods have a tedious training process, and could not cover all the challenging attributes, while the excessive number of branches also limit their inference speed.

Meanwhile, some recent works initialize the proposed RGB-T tracking models with RGB tracking models pre-trained on RGB tracking datasets. Zhang et al. \cite{mfnet} design a fusion module that can reduce the modality discrepancy and select the discriminative features, and initialize the RGB branch parameters with pre-trained DiMP parameters. Zhang et al. \cite{vtuav} integrate image-level, feature-level, and decision-level fusion into a unified framework, and use pre-trained DiMP model for parameter initialization. Based on pre-trained RGB siamese tracking parameters, Peng et al. \cite{siamivfn} propose two subnetworks to fuse visible and infrared features and determine contributions of two modalities, respectively. Hui et al. \cite{tbsi} insert multiple template-bridged interaction modules into backbone to realize robust cross-modal interaction and utilize parameters from pre-trained transformer-based RGB tracking model for full fine-tuning. By employing full-parameter fine-tuning and a meticulously crafted cross-modal fusion module, these models achieve rapid convergence even with limited training data. Consequently, they exhibit significantly superior performance compared to other RGB-T trackers without full fine-tuning. However, the huge amount of tuned parameters also bring a heavy burden for training devices, in terms of computation and storage demands, while the potential forgetting of upstream knowledge should also be fixed attention. Considering above limitations of full fine-tuning methods, some contemporary researches integrate prompt learning into RGB-T tracking to facilitate streamlined parameter-efficient fine-tuning and the proficient transference of knowledge from upstream tasks. These methods will be introduced in the next subsection.

\subsection{Visual Prompt Learning}

Prompt learning, as a parameter-efficient transferring method, is originally proposed in the natural language processing field for transferring large language models to downstream tasks. Recently, this idea is introduced to the computer vision field \cite{vpt, vpi2, dam-vp, vptg}, using a small amount of image data from downstream tasks to generate effective visual prompts for transferring pre-trained models. Jia et al. \cite{vpt} use a small number of learnable parameters as visual prompts, and prepend them to the input sequences of each layer. Bar et al. \cite{vpi2} propose to use the input-output pairs of downstream task examples and a new input image concatenated as visual prompts, to make the model learn to fill in the gaps, thus formulating this problem as an image inpainting problem. Considering that the downstream image datasets sometimes have large distribution differences, Huang et al. \cite{dam-vp} cluster the image datasets into multiple subsets with the same internal distribution, configure unique prompts for each subset, and initialize them with a meta-prompt. 

Yang et al. \cite{protrack} take the weighted addition of two modal images as visual prompts and feed the prompts into the pre-trained RGB tracking model, which is the first time that prompt learning is introduced into the multi-modal tracking task. Zhu et al. \cite{vipt} formulate the multi-modal tracking task as an RGB tracking task with an auxiliary modality input, and design a lightweight prompter to generate visual prompts from the auxiliary modality data. Hong et al. \cite{onetracker} further insert trainable adapters into foundation's transformer layer for effective prompt fine-tuning. Wu et al. \cite{untrack} combine three multimodal training sets and introduce lora techniques to learn common latent space of the additional modalities. However, to bridge the gap between the upstream and downstream tasks, these methods overemphasize the impact of the RGB modality, neglecting the dynamic modality-priority in various scenarios. Consequently, these methods fail to exploit modality-independent clues and struggle to perform robustly in intricate environments. Different with them, our prompt method effectively explores uni-modal information while adapting to scenarios through inter-modal interaction and fusion. This adaptability is achieved via flexible and diverse prompt strategies, enabling our method to excel when confronted with various challenging scenarios.

\section{Preliminaries}

\begin{figure}
    \centering
    \includegraphics[width=1\linewidth]{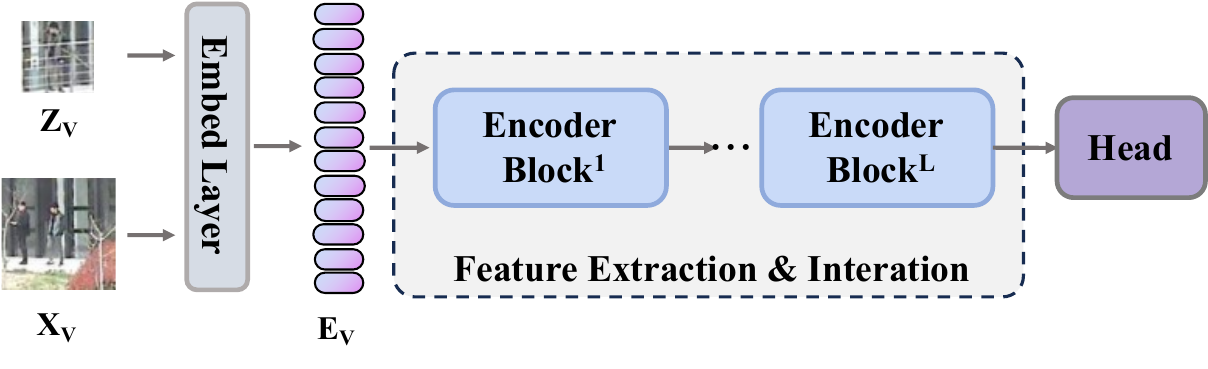}
    \caption{The tracking pipeline of the RGB-based foundation model. The foundation model is a one-stream one-stage RGB tracker based on transformer backbone. The subscript V denotes the visible modality, and the superscript i denotes the layer number of the Transformer Encoder Block.}
    \label{ostrack}
\end{figure}

Before presenting our RGB-T tracking method, we first define the RGB tracking task and briefly review the foundation model. For more details, please refer to OSTrack \cite{ostrack}.

The RGB object tracking task aims to estimate the bounding box $B$ of the target in each frame, given the bounding box $B_{0}$ of the target in the initial frame of a video sequence. The mainstream RGB tracking methods formulate this task as a similarity matching problem between the template image $ I_{V,Z}\in R^{H_{Z}\times W_{Z}\times C} $ from the initial frame and the search region $ I_{V,X}\in R^{H_{X}\times W_{X}\times C} $ from the current frame, where $V$ denotes the visible modality, and $X$ and $Z$ denote the search and template respectively. Therefore, the tracking task can be expressed as $ B=Track (I_{V,Z}, I_{V,X}) $, where $ Track() $ denotes the tracker function.

As shown in Figure \ref{ostrack}, we choose a one-stream one-stage RGB tracking model \ref{ostrack} as our foundation model. Built on ViT \cite{vit}, the method is one-stream one-stage, which means that the feature extraction and template-search interaction are performed simultaneously by a single transformer backbone. The model consists of an embedding layer, a backbone composed of $L$ transformer encoder blocks stacked together, and a head network. The embedding layer maps the input template and search images into 16$\times$16 patches $E_{V,Z}^{0}\in R^{(N_{Z})\times D}$ and $E_{V,X}^{0}\in R^{(N_{Z})\times D}$, reshapes them to $1D$ tokens and concatenates them as $E_{V}^{0}=[E_{V,Z}^{0}; E_{V,X}^{0}]\in R^{(N_{Z}+N_{X})\times D}$, where $N_{Z}$ and $N_{X}$ respectively denote the number of template tokens and search tokens and $D$ denotes the dimension of the tokens. Then, the encoder blocks $\left \{Encoder^{l}, l\in (1, L)\right \}$ perform self-attention mechanism \cite{transformer} on the concatenated tokens $E_{V}^{0}$, thus achieving joint feature extraction and interaction between the template and the search:
\begin{align}
    E_{V}^{L}=Encoder^{L}(...Encoder^{2}(Encoder^{1}(E_{V}^{0})))
\end{align}
where superscripts represent the number of layers.

Finally, the search tokens that fully interact with the template are separated from the $E_{V}^{L}$ by unconcatenation, and are reshaped by the head network into features $F_{V,X}\in R^{H_{X} \times W_{Z} \times C}$ for the target state estimation:
\begin{align}
    F_{V,Z},F_{V,X}&=reshape(unconcaten(E_{V}^{L}))\\
    &B=Head(F_{V,X})
\end{align}

\section{Methodology}

\begin{figure*}
    \centering
    \includegraphics[width=1\linewidth]{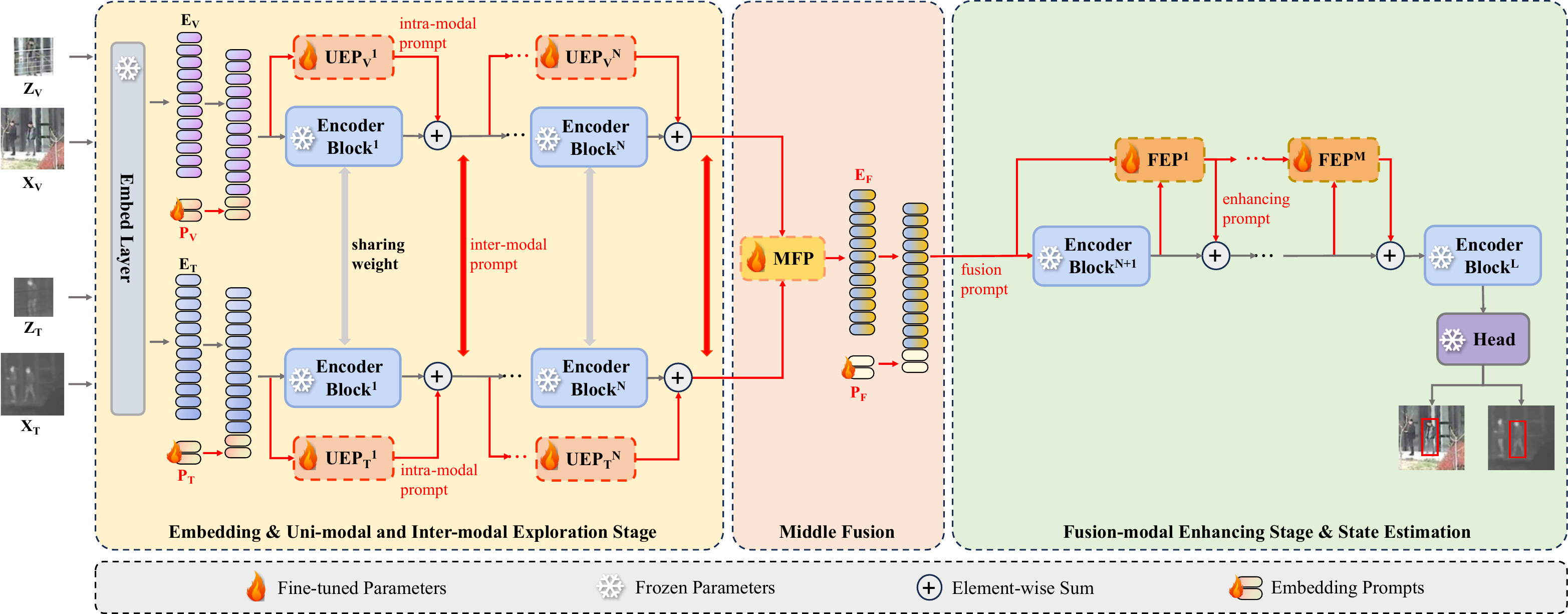}
    \caption{Pipeline of our M3PT. In this method, two modal images of the same size go through five steps: embedding, uni-modal and inter-modal exploration, middle fusion, fusion-modal enhancing, and state estimation, to obtain the predicted bounding box. Here, the L transformer encoder blocks from the upstream model are divided into two groups, for the uni-modal and inter-modal exploration and fusion-modal enhancing modeling stages respectively. The subscripts and superscripts of the modules and symbols indicate the modality and the layer number respectively. }
    \label{pipeline}
\end{figure*}

\subsection{Overview}

In this work, we propose a novel RGB-T prompt tracking method M3PT that leverages our designed middle fusion meta-framework and multi-form, multi-stage visual prompts. The overall pipeline of our method is shown in Figure \ref{pipeline}. 

We first develop a universal middle fusion meta-framework for multi-modal tracking task. In the meta-framework, the tracker's backbone consists of two parts for different stages: The first-stage backbone is a dual-stream network that extracts two uni-modal features separately; The second-stage backbone is a single-stream network that enhances the fusion-modal features. The middle fusion module is inserted between the two backbone stages.

Furthermore, based on the meta-framework, we embed the pre-trained foundation model into our tracking pipeline by using four flexible prompt strategies to achieve effective knowledge transfer. These strategies include Uni-modal and Inter-modal Exploration Prompt Strategy, Middle Fusion Prompt Strategy, Fusion-modal Enhancing Prompt Strategy and Modality-aware and Stage-aware Prompt Strategy. 

Therefore, the comprehensive pipeline of our prompt tracking approach is delineated below: (1) Following the meta-framework, the L transformer encoder blocks from the foundation model backbone are divided into two groups, containing N and M encoder blocks respectively, for feature extraction in two stages. (2) The Uni-modal and Inter-modal Exploration Prompt Strategy extends the first-group encoder blocks into a parameter-sharing dual-branch structure, and configures our designed UEP and IP in parallel with them layer by layer, to explore the modality-independent information and operate scenario-adaptive cross-modal communication, thus better adapting the encoder blocks to modeling uni-modal clues sufficiently in various intricate scenarios. (3) The Middle Fusion Prompt Strategy inserts the designed MFP after the first-stage backbone, and feeds the fusion-modal features output by MFP as visual prompts to the backbone in the later stage, to adapt it to the fusion-modal modeling. (4) The Fusion-modal Enhancing Prompt Strategy configures a lightweight prompter FEP in parallel with the second-stage backbone layer by layer, to obtain a richer fusion-modal feature representation and enhance the fusion-modal features propagated forward in the backbone. (5) The Modality-aware and Stage-aware Prompt Strategy prepends three kinds of learnable modality and stage prompts to the input of the backbones in two stages, respectively, to guide foundation model to identify the distribution characteristics of the current modality faster. (6) The head of foundation model is frozen and directly utilized to estimate the target state on the search features.

\subsection{Uni-modal and Inter-modal Exploration Prompt Strategy}

\begin{figure*}
    \centering
    \includegraphics[width=1\linewidth]{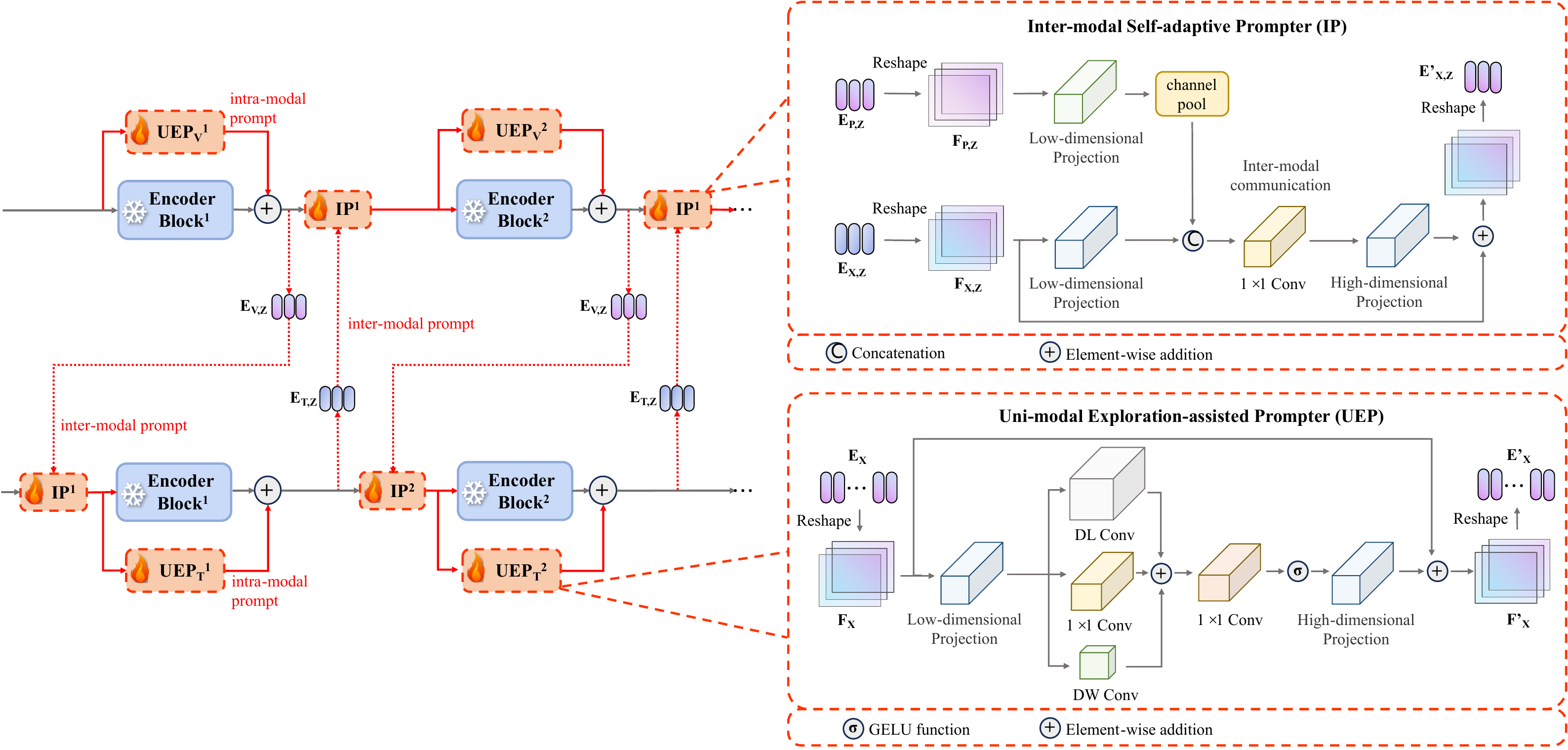}
    \caption{The pipeline of our Uni-modal and Inter-modal Exploration Prompt Strategy and overall architecture of two lightweight prompters which include Uni-modal Exploration-assisted Prompter (UEP) and Inter-modal Self-adaptive Prompter (IP).
    In the prompt strategy, the modality-independent information extracted by our designed UEP is firstly added to the output of the encoder block of the same modality as intra-modal prompts, and the prompted template tokens are further utilized by our designed IP to generate effective inter-modal scenary prompts.}
    \label{UIEP}
\end{figure*}

Previous prompt tracking methods, like ProTrack \cite{protrack}, ViPT \cite{vipt} and OneTracker \cite{onetracker}, hardly exploit enough modality-independent information and also fail to facilitate priority-aware inter-modal interaction across diverse uncontrolled environments. To tackle these problems, we propose the Uni-modal and Inter-modal Exploration Prompt Strategy, formulating two lightweight components: Uni-modal Exploration-assisted Prompter (UEP) and Inter-modal Self-adaptive Prompter (IP). Employing our prompt strategy and prompters, the first-stage backbone is directed to explore the uni-modal features while achieving scenario-adaptive inter-modal communication. The pipeline of our strategy and overall architechure of UEP and IP are shown in Figure \ref{UIEP}.

 \textbf{Uni-modal Exploration-assisted Prompter (UEP).} We design UEP for sufficient mining the discriminative patterns within each modality. As shown in Figure \ref{UIEP}, the input tokens $E_{X}$ ($X \in \left \{ V, T \right \}$) are firstly reshaped into features $F_{X}$ and mapped to a low-dimension feature space, whose purpose is reducing redundant features and trainable parameters. In the low-dimension space, we configure parallel standard 1$\times$1 convolutional layer, depth-wise convolutional layer and dilated convolutional layer to capture standard features, channel-private features, and larger receptive field local features, respectively. Through this way, modality-independent information can be thoroughly explored from various perspectives. Subsequently, an additional standard 1$ \times $1 convolutional layer with activation function GELU is employed to integrate above features. Assuming extract modality-independent features of visible modality, the entire process conducted in the low-dimensional space can be succinctly expressed as follows:
 \begin{align}
     F_{sum}&=Cv_{1}(F_{V,L})+Cv_{DL}(F_{V,L})+Cv_{DW}(F_{V,L}) \\
     &F'_{V,L} = GELU(Cv_{2}(F_{sum}))
 \end{align}
 Where subscripts $L$ and $V$ respectively denote low dimension and visible modality, $Cv_{1}()$ and $Cv_{2}()$ all denote standard 1$ \times $1 convolutional layer, $Cv_{DW}()$ and $Cv_{DL}()$ respectively denote depth-wise convolutional layer and dilated convolutional layer, and $F'$ denotes integrated features. Finally, the combined features are remapped to original high-dimension space. In addition, UEP configures residual connection. Since the feature extraction process is all performed in the low-dimension space, each UEP contains only a small number of learnable parameters. In our case, the dimensions of the low-dimension feature space of UEP is set to 8.

\textbf{Inter-modal Self-adaptive Prompter (IP).} Our Inter-modal Self-adaptive Prompter is designed to adapt the foundation model to perform robustly in priority-agnostic intricate environment. As shown in Figure \ref{UIEP}, the template tokens of prompting and prompted modality are firstly reshaped into features {$F_{P,Z}, F_{X,Z}$} (Here $P$ and $X$ respectively denote prompting and prompted modality, and $Z$ denotes template.) and mapped to a low-dimension feature space. The reason for using template features as inter-modal prompts is that the templates contain less background noise and more clues about the target appearance. Then, we employ average channel-pooling to capture cross-channel clues of the prompting modality features $F_{P,Z,L}$ ($L$ denotes low dimension.) and take the one-channel feature as a channel-level inter-modal prompt $P_{P, inter}$. To elaborate, we append the prompt $P_{P, inter}$ as an as an auxiliary channel to the prompted features $F_{X,Z,L}$ and amalgamate them via a standard 1$\times$1 convolution layer. Assuming we designate visible modality as the prompting modality and thermal infrared modality as the prompted modality, the entire process conducted in the low-dimensional space can be succinctly expressed as follows: 
\begin{align}
    &P_{V,inter} = ACP(F_{V,Z,L}) \\
    F'_{T,Z,L} &= Cv(concaten([P_{V,inter},F_{T,Z,L}]))
\end{align}
where $ACP()$ represents average channel pooling, $Cv()$ denotes a standard 1$\times$1 convolution layer and $F'$ denotes combined prompted features. Finally, a high-dimension projection and a residual connection are configured, which is similar to UEP. In our case, the dimensions of the low-dimension feature space of IP is set to 8. \textbf{Why we choose this channel-level prompt approach?} It is noteworthy that existing prompt methods primarily sum the two modalities with employ constant global weights or element-wise weights (acquired through learning), to achieve inter-modal communication. While the latter methods generally surpass the former in effectiveness, static weights consistently fall short in adapting the model proficiently to the dynamic modality dominance in complex, open-world scenarios. Different with them, our prompter further learns how to infuse the critical information of the prompting modality, captured via channel pooling, into the prompted modality, rather than indiscriminately applying fixed weights to both modalities. Consequently, our strategy demonstrates enhanced adaptability when confronted with complex scenarios characterized by variable modality priority.

The proposed Uni-modal and Inter-modal Exploration Prompt Strategy utilize modality-private UEP and modality-sharing IP to adapt first stage backbone to thoroughly explore the coupled feature space of the two modalities as illustrated in Figure \ref{UIEP}. Specifically, the prompt pipeline could be described as follows: (1) For uni-modal prompt process, the modality-independent features extracted by UEP is added as intra-modal prompts to the modality-shared features output by Encoder Block, thus providing a richer uni-modal representation for the next backbone layer. Taking the visible branch as an example, this prompt process can be expressed as:
\begin{align}
    &P_{V,intra}^{n}=UEP_{V}^{n}(H_{V}^{n}) \\
    E_{V}^{n}&=Encoder^{n}(H_{V}^{n})+P_{V,intra}^{n}
\end{align}
where superscript $n$ denotes layer number, $H$ denotes input tokens of an encoder block, and $P$ denotes the prompts generated by designed prompter.
(2) For inter-modal prompt process, current and another modality branches are respectively taken as prompting and prompted features, the template tokens of them are input to IP, realizing inter-modal prompting. Notably, here we adopt an asymmetric cross-layer manner for inter-modal prompting, which could formulated as follows:
\begin{align}
    H_{T,Z}^{n}&=IP^{n}(E_{T,Z}^{n-1},E_{V,Z}^{n})\\
    H_{V,Z}^{n+1}&=IP^{n}(E_{T,Z}^{n},E_{V,Z}^{n})
\end{align}

\subsection{Middle Fusion Prompt Strategy}

\begin{figure*}
    \centering
    \includegraphics[width=1\linewidth]{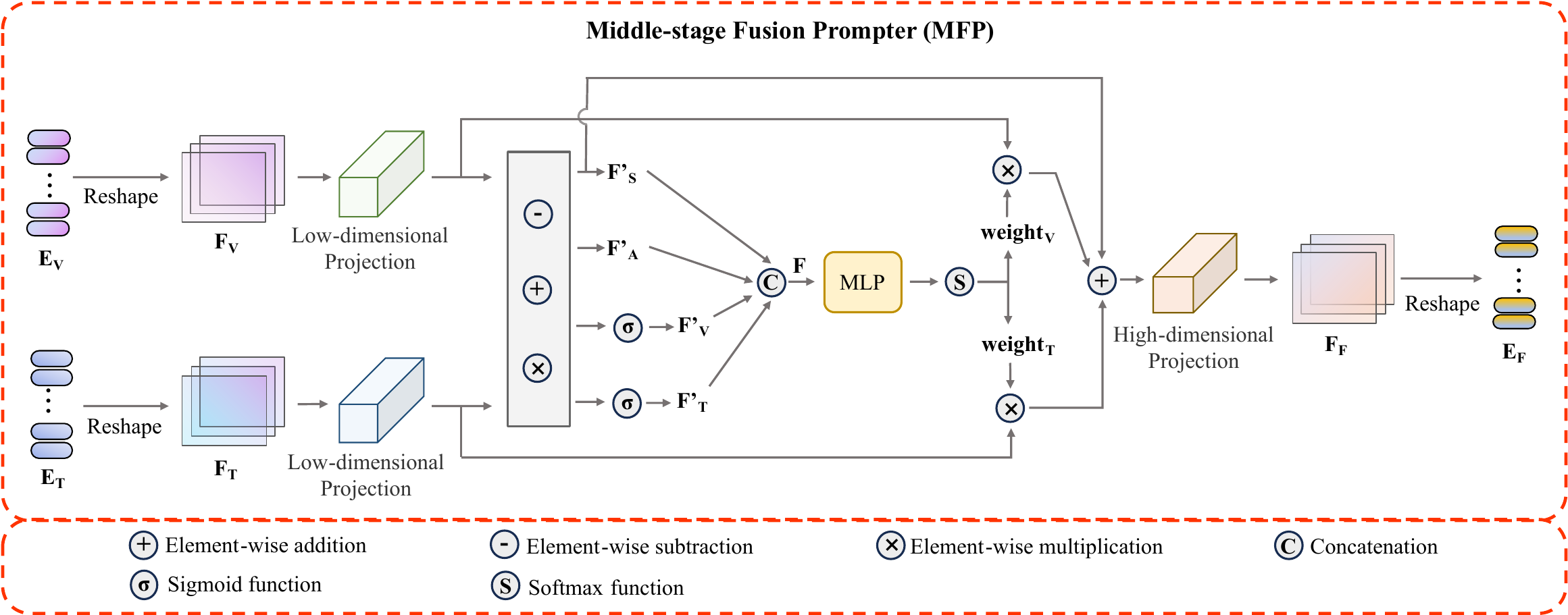}
    \caption{Overall structure of our designed Middle Fusion Prompter (MFP).}
    \label{MFP}
\end{figure*}

To achieve full integration of the uni-modal features extracted by the first-stage backbone and provide reliable task prompts for the subsequent backbone, we propose the Middle Fusion Prompt Strategy that utilizes our designed Middle Fusion Prompter (MFP) to perform active selection of discriminative features and self-adaptive fusion, and take the fusion-modal features as natural visual prompts for the next stage.

\textbf{Middle Fusion Prompter (MFP).} MFP is shown in Figure \ref{MFP}. The input feature tokens of the two modalities $E_{V}$ and $E_{T}$ are first reshaped into features $F_{V}$ and $F_{T}$, and then mapped to low-dimension feature space by two modality-private standard 1$ \times $1 convolutional layers, respectively. Here, we set the dimensions of the low-dimension feature space of the MFP to 16. Then, inspired by \cite{cmd}, we separate the modality-shared and modality-independent information extracted at each element: 
\begin{align}
    F'_{S}&=(F_{V,L}\times F_{T,L}) \\
    F'_{A}&=(F_{V,L}+F_{T,L}) \\
    F'_{V}=&Sigmoid(F_{V,L}-F_{T,L}) \\
    F'_{T}=&Sigmoid(F_{T,L}-F_{V,L})
\end{align}
where $F_{V,L}$ and $F_{T,L}$ represent the visible light and infrared features mapped to the low-dimension space, respectively, $F'_{S}$ represents the modality-sharing features, $F'_{A}$ represents the overall features, $F'_{V}$ and $F'_{T}$ represent the visible light and infrared modality-independent features, respectively, $\times$, $+$, and $-$ represent element-wise multiplication, addition, and subtraction, respectively, and sigmoid operation is used to enhance the local modality-independent features. After that, we concatenate these features in the following order and generate adaptive discriminative filtering weights based on them:
\begin{align}
    F=concaten(F'_{V}, F'_{S}, F'_{A}, F'_{S}, F'_{T})\\
    [weight_{V},weight_{T}]=Softmax(fc(F))
\end{align}
where $fc()$ denotes fully connected layer and $concaten()$ denotes concatenation. And the adaptive discriminative fusion process is expressed as follows:
\begin{align}
    F_{F,L}=F'_{S}+ F_{V,L}\times weight_{V}+ F_{T,L}\times weight_{T}
\end{align}
It is noteworthy that we add $F'_{S}$ to the weighted fusion features, considering that certain useful information existing at the low-weight positions in the discriminative fusion process will be suppressed and these low-weight information mostly belong to the modality-sharing information. Previous weighted fusion methods have ignored it. Therefore, we ensure the integrity of the effective information by adding $F'_{S}$. Finally, the fused features are remapped to the original dimension and reshaped into tokens. 

 We propose the Middle Fusion Prompt Strategy that inserts MFP after the first-stage backbone, which contains N Encoder Blocks. Unlike \cite{protrack} and \cite{vipt}, which only perform image-level weighted fusion of the two modal features, we perform adaptive and discriminative selection and complementary fusion of the two modal features that have been fully explored and modeled, and we retain as much useful information as possible. Therefore, the obtained fusion-modal features can serve as reliable fusion-modal prompts injected into the second-stage upstream backbone, making it better adapt to the fusion-modal modeling task.

Furthermore, to the best of our knowledge, this is the first time that the middle fusion meta-framework is introduced to the RGB-T tracking task. Compared to the image-level fusion framework, our meta-framework can help the tracker more effectively explore the independent information of the two modalities; compared to the feature-level fusion and decision-level fusion framework, our meta-framework can help the tracker effectively compress the redundant modeling computation and reduce the model parameters. Our prompt method provide a typical application instance. While keeping the total amount of backbone parameters from the pre-trained model unchanged, by flexibly adjusting the fusion location, the proportion allocation of the two-stage backbone parameters could be freely changed, balancing the performance and efficiency for diverse application requirements.

\subsection{Fusion-modal Enhancing Prompt Strategy}

To further narrow the gap between the second-stage modeling and the upstream modeling, we propose a Fusion-modal Enhancing Prompt Strategy, which introduces the fusion-modal enhancing prompter FEP to provide enhanced prompts for the fusion-modal features. Here, FEP adopts the same structure as MCP in \cite{vipt}, and is also configured in parallel with encoder blocks layer by layer, as shown in Figure 3. But unlike \cite{vipt}, our second-stage backbone input is a single-stream fusion features, which serves as one of the input streams of the first-layer FEP, and the output of the first-layer Encoder Block serves as another input stream of the first-layer FEP. And the output of FEP serves as the visual cue for self-enhancement and one of the input streams of the next-layer FEP, which can be expressed as:
\begin{align}
  P^{m}=&FEP^{m-N}(H^{m},Encoder^{m}(H^{m})),m=N+1 \\
  P^{m}=&FEP^{m-N}(P^{m-1},Encoder^{m}(H^{m})),m>N+1 \\
  &H^{m+1}=Encoder^{m}(H^{m})+P^{m},m>N+1
\end{align}
where superscript denotes layer number, $H$ denotes input tokens of an encoder block, and $P$ denotes the prompts generated by designed prompter.

By using this quasi-residual connection method, the local key information of the fusion-modality is enhanced and injected into the next layer, providing a more rich and reliable fusion-modal feature representation, which makes the subsequent upstream backbone adapt to the fusion-modal modeling.

\subsection{Modality-aware And Stage-aware Prompt Strategy}

We intuitively assume that each modality contains some fixed patterns such as low-level distribution characteristics, which can be learned and stored to guide the backbone to quickly and familiarize themselves with the current modality and modeling task. Therefore, we propose Modality-aware and Stage-aware Prompt Strategy that stores the fixed patterns of the visible, thermal, and fusion modalities by three kinds of learnable prompts. Inspired by \cite{vpt}, we prepend these prompts to the input of corresponding encoder blocks, to provide upstream backbone with guidance of current modality and stage, as shown in Figure \ref{pipeline}. 

These prompts only participate in the feature extraction of upstream backbone. Namely, the prompts are separated from total tokens at the input of all prompter branches, and are concatenated with the prompted feature tokens after completing other prompting processes. For an instance, after MFP completes the modality feature fusion, the two uni-modal prompts are added to the fusion-modal prompts which then are prepended to the input of second-stage backbone. Since they do not contain positional encoding, these prompts can be inserted into any position of the feature tokens. In our case, we set the token number of the learnable prompts to 2.

\subsection{Head, Loss and Training}

The head network parameters of the base model are frozen and directly used in our framework to estimate the current state of the target. The loss function during training is also the same as the base model, including the weighted classification loss, GIoU loss, and L1 loss, whose formulas are as follows:
\begin{align}
    L=L_{cls}+\lambda_{iou} L_{giou}+\lambda_{L1} L_{L1}
\end{align}
Where $\lambda_{giou}$ and $\lambda_{L1}$ are exactly the same as the training settings of the foundation model.

\section{Experiment}

\begin{figure*}
    \centering
    \includegraphics[width=1\linewidth]{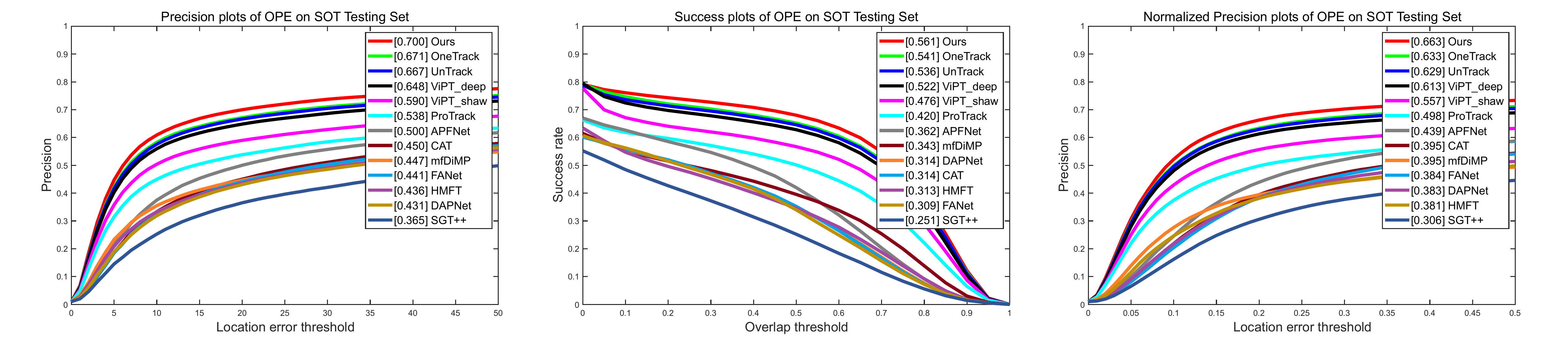}
    \caption{The evaluation curves of metrics PR, SR and NPR on LasHer.}
    \label{lasher}
\end{figure*}

\begin{figure*}
    \centering
    \includegraphics[width=0.83\linewidth]{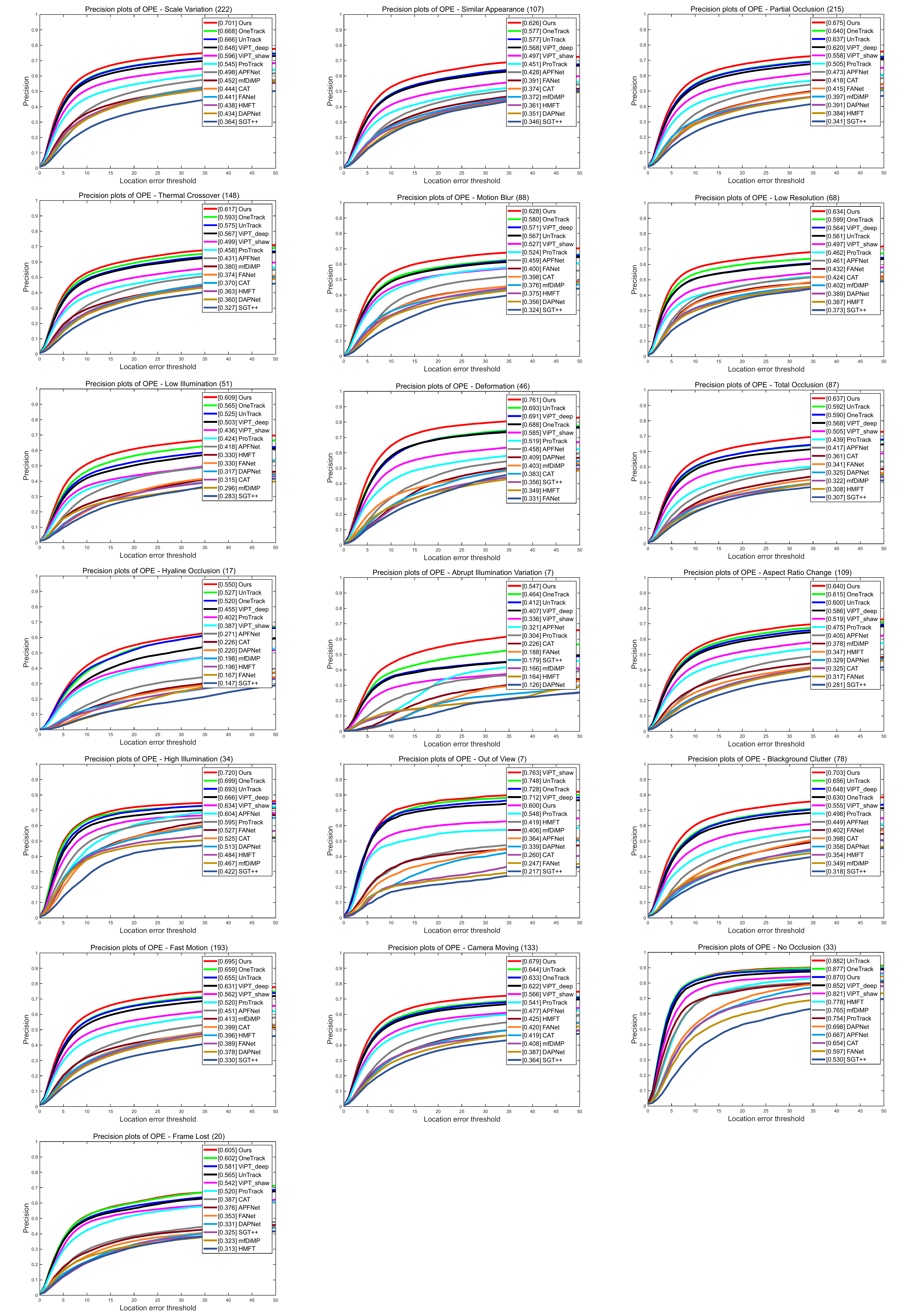}
    \caption{The evaluation curves of MPR for 19 challenge attributes on LasHer.}
    \label{lasher_attr}
\end{figure*}

\begin{figure}
    \centering
    \includegraphics[width=1\linewidth]{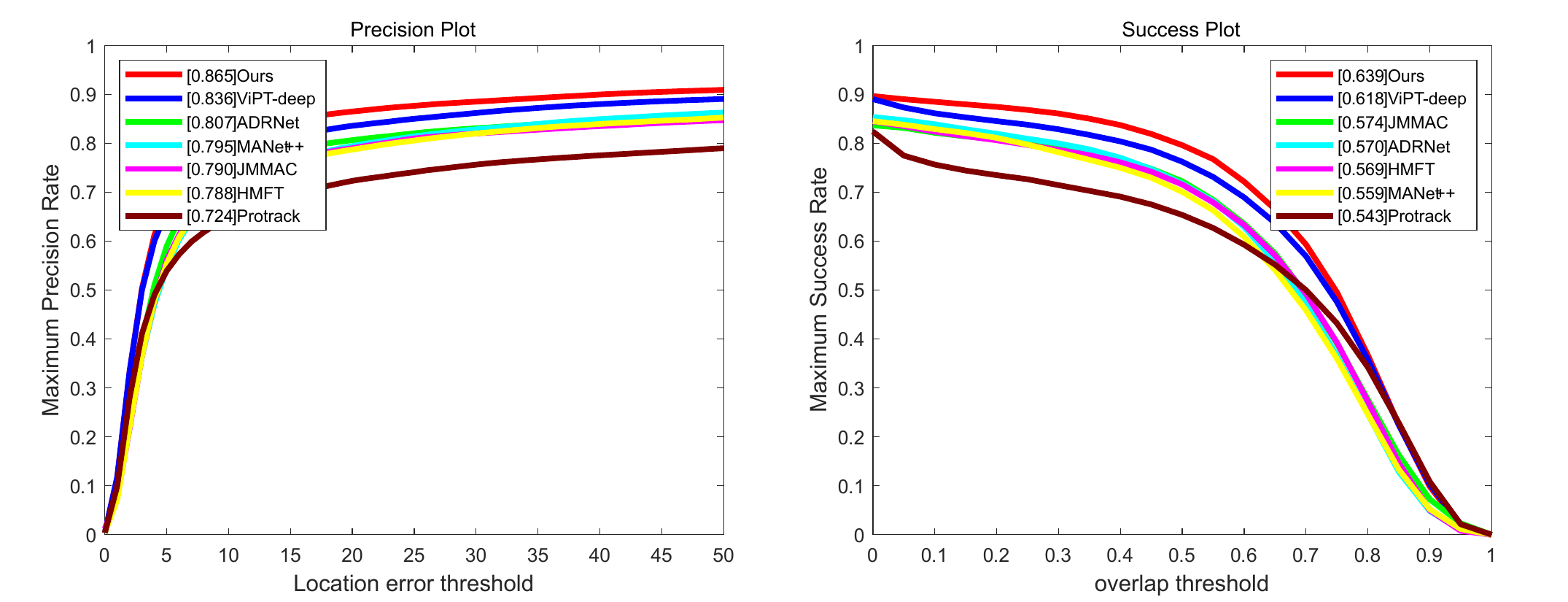}
    \caption{The evaluation curves of metrics MPR and MSR on RGBT234.}
    \label{rgbt234}
\end{figure}

\begin{figure*}
    \centering
    \includegraphics[width=1\linewidth]{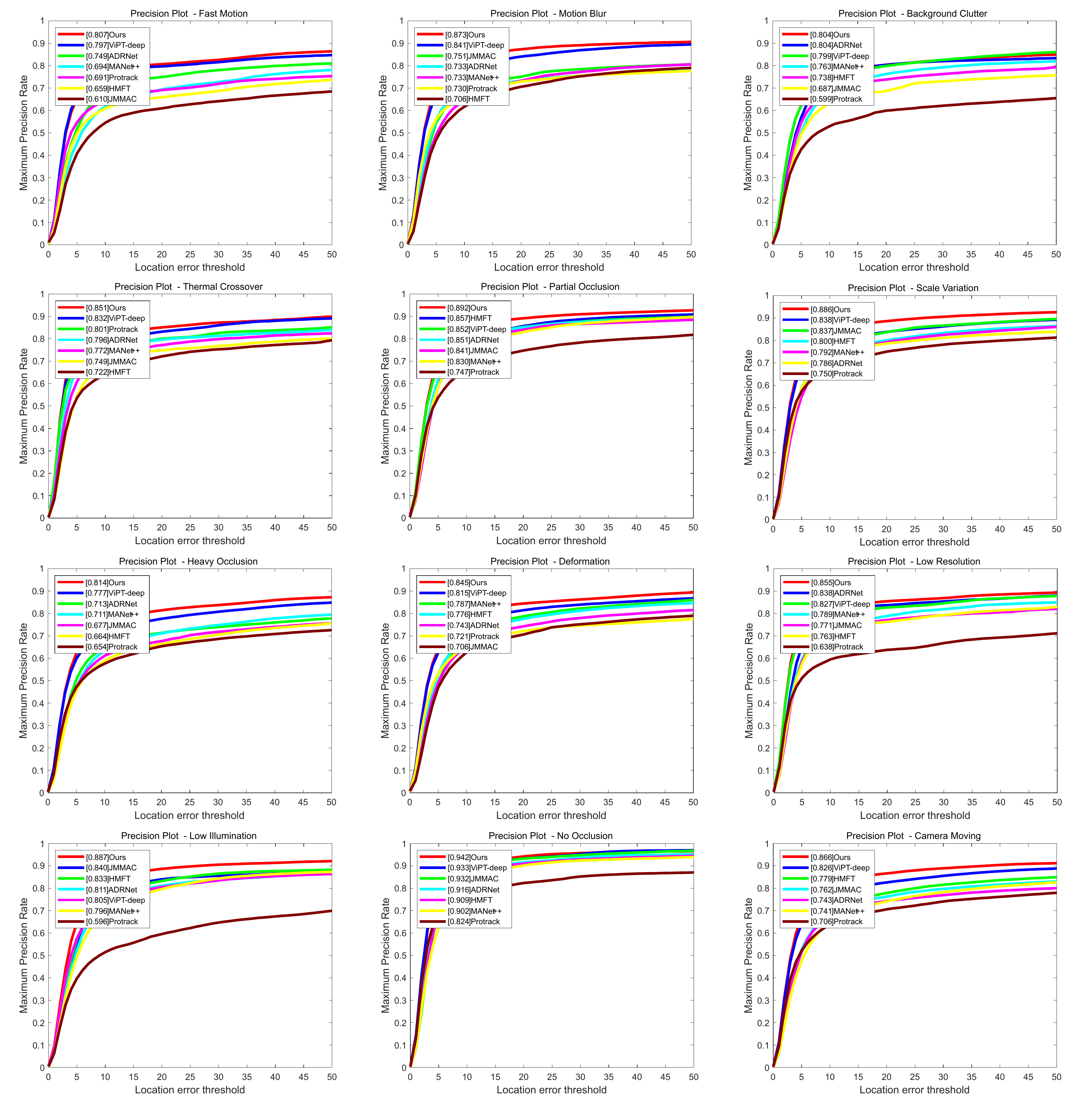}
    \caption{The evaluation curves of MPR for 12 challenge attributes on RGBT234.}
    \label{rgbt234_attr}
\end{figure*}

\begin{figure}
    \centering
    \includegraphics[width=1\linewidth]{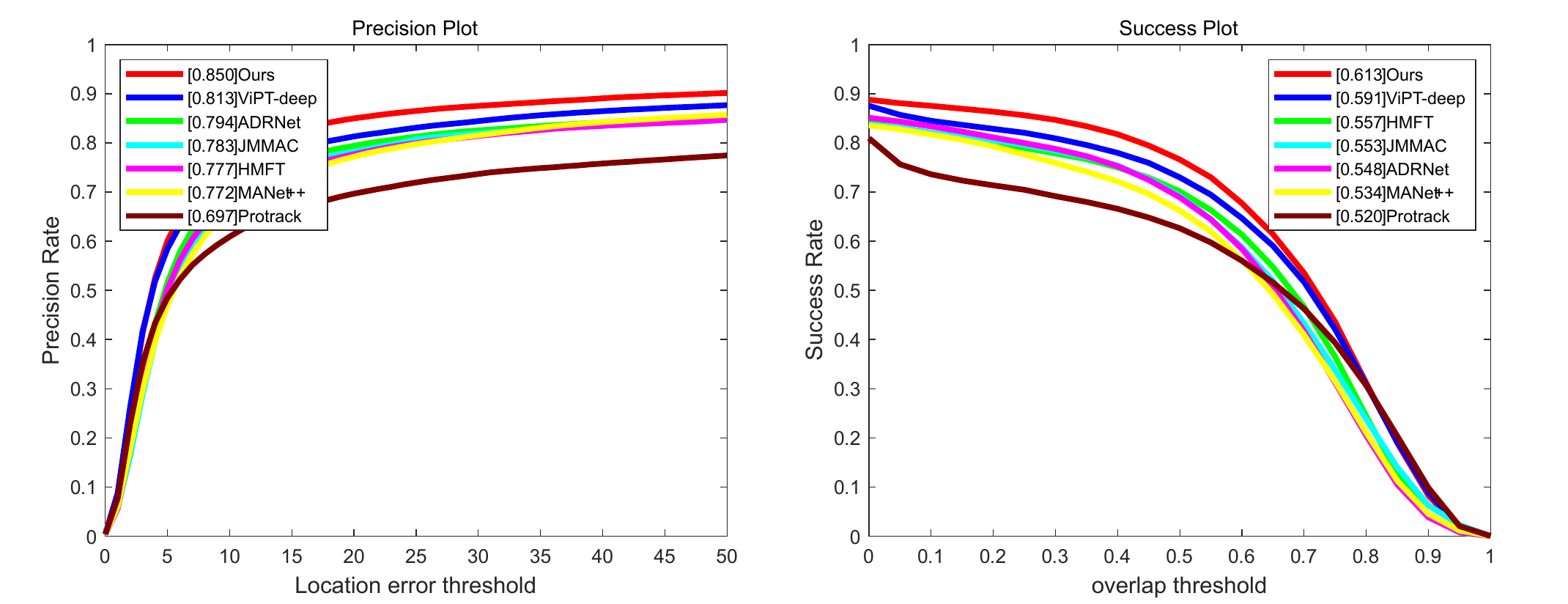}
    \caption{The evaluation curves of metrics PR and SR on RGBT210.}
    \label{rgbt210}
\end{figure}

\begin{figure}
    \centering
    \includegraphics[width=1\linewidth]{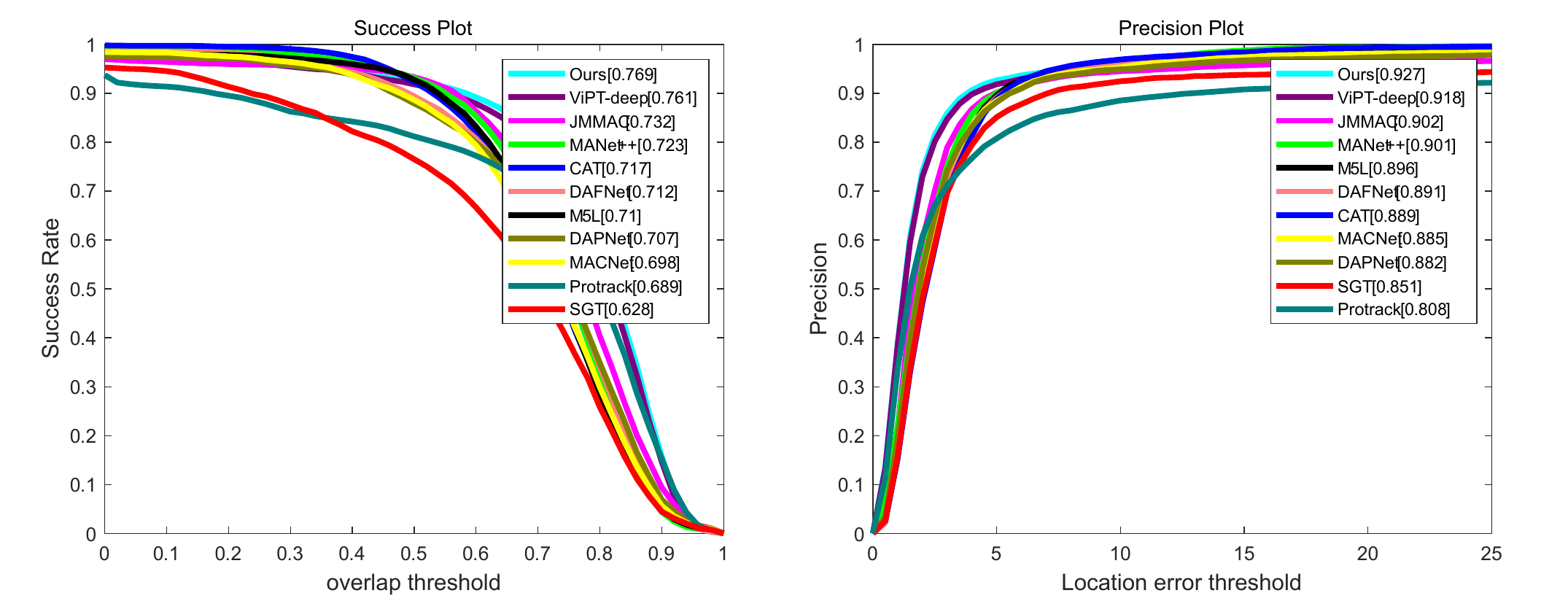}
    \caption{The evaluation curves of metrics PR and SR on GTOT.}
    \label{gtot}
\end{figure}

\begin{figure}
    \centering
    \includegraphics[width=1\linewidth]{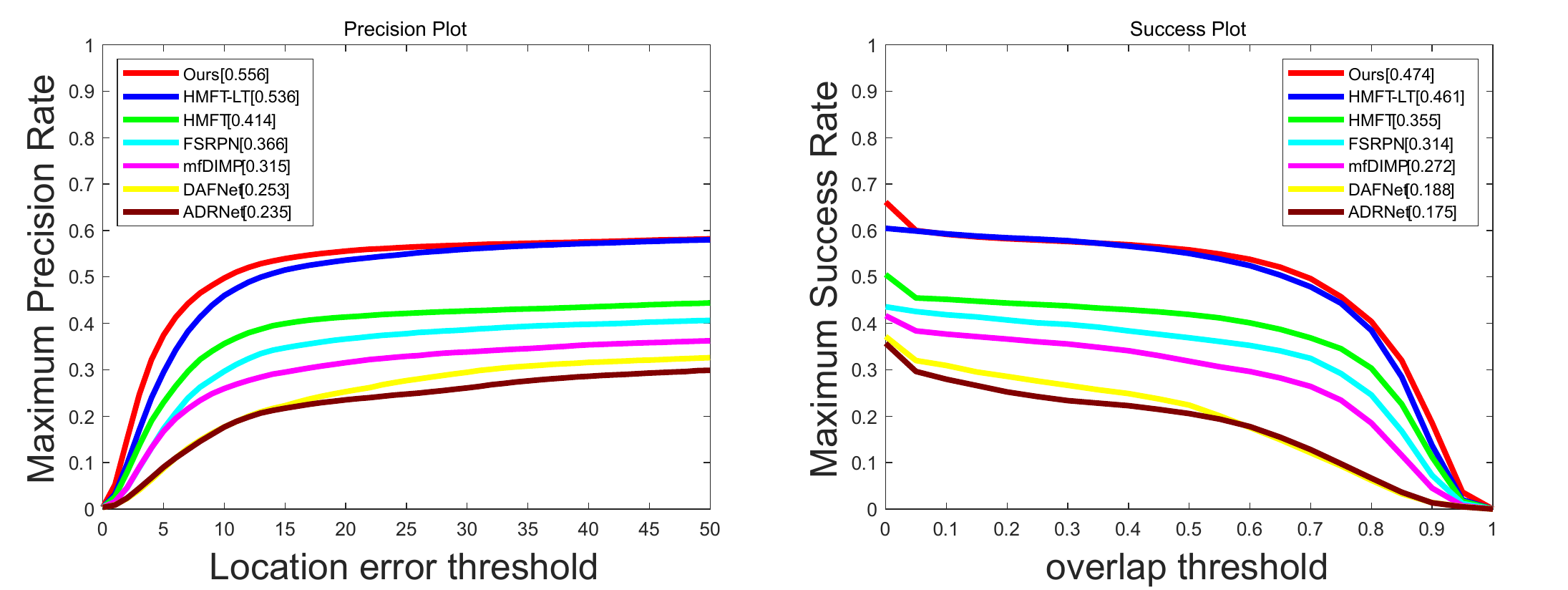}
    \caption{The evaluation curves of metrics MPR and MSR on long-term test subset of VTUAV.}
    \label{vtuavlt}
\end{figure}

\begin{figure}
    \centering
    \includegraphics[width=1\linewidth]{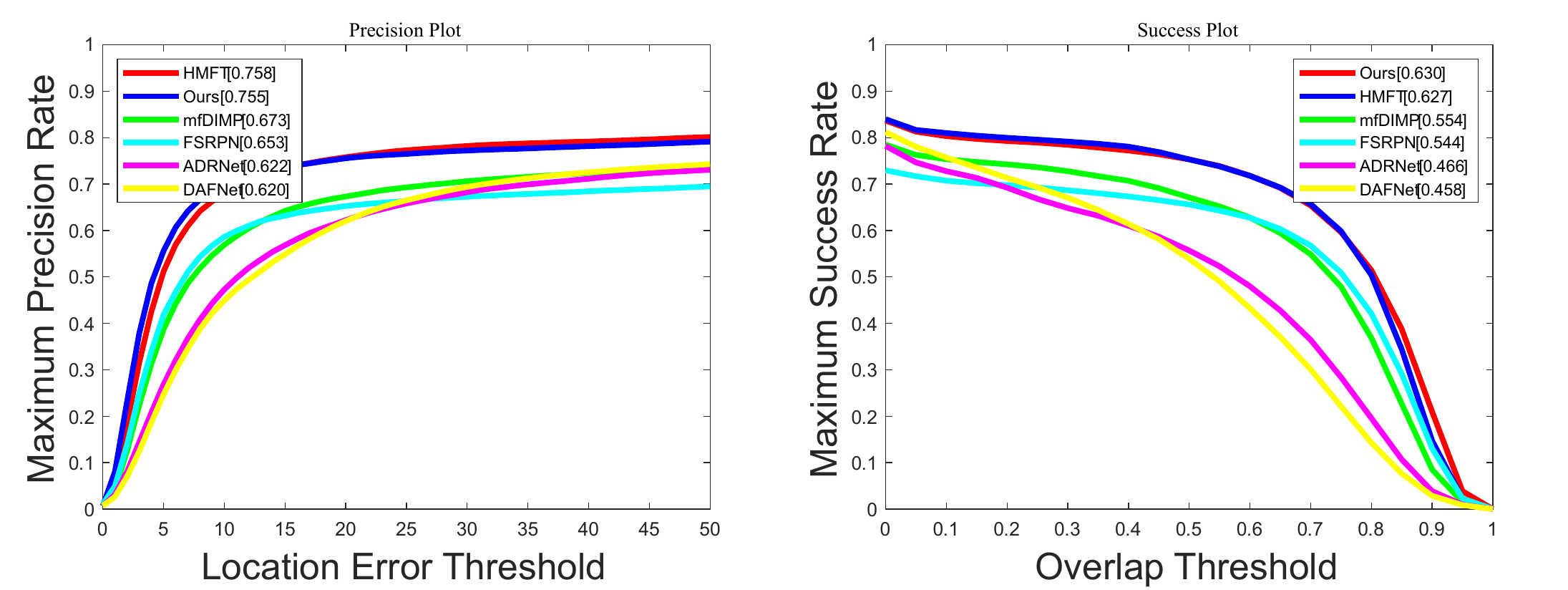}
    \caption{The evaluation curves of metrics MPR and MSR on short-term test subset of VTUAV.}
    \label{vtuavst}
\end{figure}

\begin{figure*}
    \centering
    \includegraphics[width=1\linewidth]{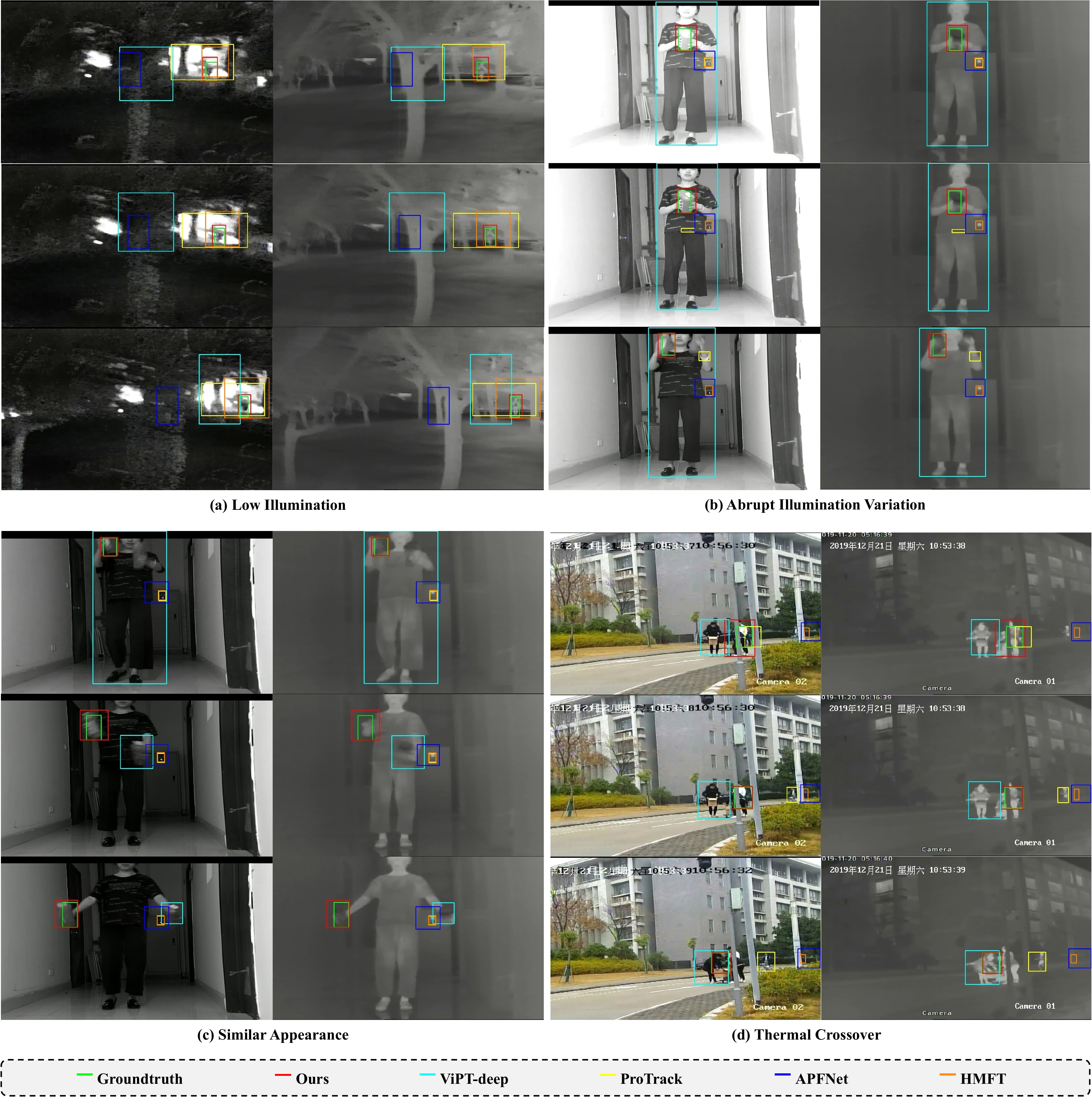}
    \caption{The qualitative evaluation results on the four most typical and challenging scenarios sampled. The green, red, cyan, yellow, blue, and orange bounding boxes are from the ground truth, our method, ViPT-deep, ProTrack, APFNet, and HMFT, respectively.}
    \label{all}
\end{figure*}

\subsection{Implementation Details}

\textbf{Foundation Model.} We choose the pre-trained OSTrack-256 model as the foundation model and fine-tune our M3PT in RGB-T tracking task.

\textbf{Training.} For fair evaluation and comparison, we train our model on LasHer training set and whole LasHer, respectively. During training, we freeze the foundation model and only train prompt parameters. The batch size, number of epochs and sample pairs of each epoch are set to 32, 60 and 60000, respectively. We use AdamW optimizer with a weight decay rate of 0.0001 and an initial learning rate of 0.0004, which decays to 0.1 times the original value after 48 epochs. The training process is end-to-end. The template and search image sizes are set to 128×128 and 256×256, respectively. The block number of the first-stage backbone is set to 10. To regulate the number of fine-tuning parameters during the fine-tuning process, we selectively employ UEP only for the 2nd, 5th and 8th Encoders. We conduct our training on four Nvidia RTX 3090 GPUs, using Python and PyTorch.

\textbf{Inference and Evaluation.} The model trained on Lasher training set and whole LasHer are evaluated on LasHer testing set and other benchmarks, respectively. There are not online parameter or template update operated during the inference.

\subsection{Experimental Benchmarks}

To sufficiently verify the effectiveness of our method, we evaluate its performance on six challenging and public RGB-T tracking benchmarks: LasHer \cite{lasher}, RGBT234 \cite{rgbt234}, RGBT210 \cite{rgbt210}, VTUAV \cite{vtuav}, GTOT \cite{gtot} and VOT2019RGB-TIR \cite{vot2019rgbt}, which encompass a diverse range of tracking scenarios, challenging attributes, and object categories, spanning rich long and short video sequences. We will briefly describe them in the following paragraphs.

\textbf{LasHer:} LasHer is the largest-scale RGB-T tracking benchmark up to now, containing 1224 highly aligned video sequence pairs and 734.8K image pairs. The benchmark covers 32 common object categories and 19 typical challenge attributes, posing great challenges for current RGB-T tracking algorithms. The official evaluation metrics of LasHeR are Precision Rate (PR), Normalized Precision Rate (NPR), and Success Rate (SR).

\textbf{RGBT210:} RGBT210 consists of 210 video sequence pairs and 210K image pairs, with the longest video sequence having 8K frames. The official evaluation metrics of RGBT210 are PR and SR.

\textbf{RGBT234:} RGBT234 is the extension version of RGBT210, containing 234 RGB-T video sequence pairs and 234K image pairs. The benchmark includes 12 common challenge attributes. The official evaluation metrics of RGBT234 are Maximum Precision Rate (MPR) and Maximum Success Rate (MSR).

\textbf{GTOT:} GTOT is the first large-scale benchmark in RGB-T tracking field. 50 pairs of annotated RGB-T sequences are contained in its dataset. PR and SR are employed as the official evaluation metrics of GTOT.

\textbf{VTUAV:} VTUAV is a large-scale benchmark, comprising 500 video sequence pairs and 1700K image pairs, with high resolution. It is divided into long-term subset and short-term subset, each containing 250 RGB-T video sequence pairs. VTUAV is the first RGB-T tracking benchmark that formally proposes the long-term tracking task. The official evaluation metrics employed on VTUAV are MPR and MSR.

\textbf{VOT2019RGB-TIR:} VOT2019RGB-TIR benchmark is introduced in the RGB-T tracking channel of VOT2019 \cite{vot2019rgbt} tacking challenge and contains 60 pairs of RGB-T sequences. The evaluation metrics employed on the benchmark are Expected Average Overlap (EAO), Accuracy (AUC), and Robustness (R).

\subsection{Evaluation Results on LasHer}

\begin{table}[h]
\caption{The overall performance on LasHer. The best two results among each type of methods are shown in \textcolor{red}{\textbf{red}} and \textcolor{blue}{\textbf{blue}} fonts, respectively.}\label{lasher_result}
\begin{tabular}{@{}lllll@{}}
\toprule
 Method & Type & PR$\uparrow $ & NPR$\uparrow $ & SR$\uparrow $ \\
 \midrule
 M3PT (Ours) & Prompt Fine-tune&\textcolor{red}{\textbf{70.0}}&\textcolor{red}{\textbf{66.3}}&\textcolor{red}{\textbf{56.1}}\\
OneTrack \cite{onetracker}&Prompt Fine-tune&\textcolor{blue}{\textbf{67.1}}&\textcolor{blue}{\textbf{63.3}}&\textcolor{blue}{\textbf{54.1}}\\
UnTrack \cite{untrack}&Prompt Fine-tune&66.7&62.9&53.6\\
ViPT-deep \cite{vipt}&Prompt Fine-tune&64.8&61.3&52.2\\
ViPT-shaw \cite{vipt}&Prompt Fine-tune&59.0&55.7&47.6\\
ProTrack \cite{protrack}&Prompt Fine-tune&53.8&49.8&42.0\\
\midrule
TBSI \cite{tbsi}&Full Fine-tune&\textcolor{red}{\textbf{69.2}}&\textcolor{red}{\textbf{65.7}}&\textcolor{red}{\textbf{55.6}}\\
ECMD \cite{cmd}&Full Fine-tune&59.0&54.6&46.4\\
DRGCNet \cite{drgcnet}&Full Fine-tune&48.3&42.3&33.8\\
MFNet \cite{mfnet}&Full Fine-tune&\textcolor{blue}{\textbf{59.7}}&\textcolor{blue}{\textbf{55.4}}&\textcolor{blue}{\textbf{46.7}}\\
APFNet \cite{apfnet} &Full Fine-tune &50.0&43.9&36.2\\
HMFT \cite{vtuav}&Full Fine-tune&43.6&38.1&31.3\\
mfDiMP \cite{mfdimp}&Full Fine-tune&44.7&39.5&34.3\\
SGT++ \cite{rgbt234}&Full Fine-tune&36.5&30.6&25.1\\
\midrule
PLASSO \cite{plasso}&Other&\textcolor{red}{\textbf{54.7}}&-&\textcolor{red}{\textbf{47.2}}\\
SiamMLAA \cite{siammlaa}&Other&\textcolor{blue}{\textbf{53.8}}&-&\textcolor{blue}{\textbf{43.1}}\\
MANet++ \cite{manet++}&Other&46.7&\textcolor{red}{\textbf{40.8}}&31.7\\
CAT \cite{cat}&Other&45.0&\textcolor{blue}{\textbf{39.5}}&31.4\\
FANet \cite{fanet}&Other&44.1&38.4&30.9\\
DAPNet \cite{dapnet}&Other&43.1&38.3&31.4\\
\bottomrule
\end{tabular}
\end{table}

We evaluate the performance of our proposed M3PT on LasHer, and compare it with other 19 state-of-the-art RGB-T tracking methods. To better contrast methods of the same type as well as those of differing types, we carefully divide all 20 methods into three categories: the first category consists of 6 lately excellent methods that utilize pre-trained RGB tracking models for parameter-efficient prompt fine-tuning; the second category comprises 8 cutting-edge methods that take pre-trained RGB tracking models for full-parameter fine-tuning; the third category includes 6 recent methods that do not use pre-trained upstream parameters.

Table \ref{lasher_result} shows the evaluation results of all methods on LasHer. (1) When compared with similar methods, to Our proposed M3PT achieves PR, NPR, and SR scores of 70.0, 66.3, and 56.1 respectively, which are \textbf{5.2}, \textbf{5.0}, and \textbf{3.9} percentage points higher than our baseline ViPT-deep \cite{vipt} respectively and wins the first place ranks among all existing prompt fine-tuning methods. This demonstrates the superiority of our proposed method over the previous prompt methods. (2) Compared with other two types of methods, our method not only outperforms all state-of-the-art methods without tuning, but also surprisingly beats the best full fine-tuning method TBSI \cite{tbsi}. Remarkably, the tuned parameters of our prompt method are \textbf{less than 1$\%$} of the sota full fine-tuning methods, which will be further analysed in subsection 5.9. In addition, we also use the LasHer toolkit to plot the PR, NPR, and SR evaluation curves of our method and other 12 open-source state-of-the-art methods, as shown in Figure \ref{lasher}. Our method shows great competitiveness under all thresholds of each evaluation curve. 

LasHer provides fine annotation of 19 classic attributes, encompassing a general attribute—No Occlusion (NO)—and 18 intricate attributes: Partial Occlusion (PO), Total Occlusion (TO), Hyaline Occlusion (HO), Low Illumination (LI), High Illumination (HI), Abrupt Illumination Variation (AIV), Out-of-View (OV), Low Resolution (LR), Deformation (DEF), Background Clutter (BC), Similar Appearance (SA), Thermal Crossover (TC), Motion Blur (MB), Camera Moving (CM), Frame Lost (FL), Fast Motion (FM), Scale Variation (SV), and Aspect Ratio Change (ARC). These attributes can cover most intricate scenarios in the real world. To evaluate our proposed method in challenging scenarios, we use official toolkit to plot attribute-wise PR evaluation curves of our method and 12 other open-source methods, as shown in Figure \ref{lasher_attr}. The vital conclusions could be summarized as follows: (1) In general case NO, because most sota trackers could easily track the target without any challenging disruptions, our method does not exhibit a pronounced competitive advantage against latest sota OneTrack and UnTrack. (2) In challenging cases, the advantage and robustness of our M3PT is fully demonstrated through surpassing all sota methods in \textbf{17 of the 18} challenging attributes (PO, FL, TC, SV, SA, MB, LR, LI, HI, DEF, CM, BC, ARC, AIV, TO, HO, FM), which include not only typical challenge attributes of visual tracking (TO, HO, FL, SV, PO, SA, MB, ARC, LR, CM, FM, BC, DEF), but also the extremely challenge attributes of visible modality (LI, HI, AIV) and thermal infrared modality (TC). The results proves that our method can indeed transfer upstream knowledge effectively and release great potential of multi-modal prompt tracking by uni-model exploration and scenario-adaptive fusion. (3) In the cases of OV, our tracker does not get leading performance. The result may be attributed to the amplification of location error in two branches and inter-modal fusion module, when images of both modalities all lose the target appearance. We think extra clues like motion prompts might be useful to solve this problem while maintaining our great performance on other challenging attributes.

\subsection{Evaluation Results on RGBT234}

\begin{table}[h]
\caption{The overall performance on RGBT234. The best two results among each type of methods are shown in \textcolor{red}{\textbf{red}} and \textcolor{blue}{\textbf{blue}} fonts, respectively.}\label{rgbt234_result}
\begin{tabular}{@{}llll@{}}
\toprule
 Method & Type & MPR$\uparrow $ & MSR$\uparrow $ \\
 \midrule
M3PT (Ours) & Prompt Fine-tune & \textcolor{red}{\textbf{86.5}} & \textcolor{blue}{\textbf{63.9}} \\
OneTrack \cite{onetracker}&Prompt Fine-tune&\textcolor{blue}{\textbf{85.7}}&\textcolor{red}{\textbf{64.2}}\\
UnTrack \cite{untrack}&Prompt Fine-tune&83.7&61.8\\
ViPT-deep \cite{vipt}&Prompt Fine-tune&83.6&61.8\\
ProTrack \cite{protrack}&Prompt Fine-tune&72.4&54.3\\
\midrule
USTrack \cite{ustrack} &Full Fine-tune& \textcolor{red}{\textbf{87.4}} & \textcolor{red}{\textbf{65.8}} \\
TBSI \cite{tbsi} &Full Fine-tune& \textcolor{blue}{\textbf{87.1}} & \textcolor{blue}{\textbf{63.7}} \\
ECMD \cite{cmd} &Full Fine-tune& 82.4 & 58.4 \\
SiamIVFN \cite{siamivfn} &Full Fine-tune& 81.1 & 63.2 \\
DRGCNet \cite{drgcnet} &Full Fine-tune& 82.5 & 58.1 \\
MFNet \cite{mfnet} &Full Fine-tune& 84.4 & 60.1\\
DMCNet \cite{dmcnet} &Full Fine-tune & 83.9 & 59.3  \\
APFNet \cite{apfnet}  &Full Fine-tune & 82.7 & 57.9 \\
 HMFT \cite{vtuav} &Full Fine-tune & 78.8 & 56.9 \\
\midrule
SiamAFTS \cite{siamafts} &Other& \textcolor{red}{\textbf{89.0}} & \textcolor{blue}{\textbf{60.2}} \\
 PLASSO \cite{plasso} & Other & \textcolor{blue}{\textbf{82.7}} & 59.2 \\
 SiamMLAA \cite{siammlaa} &Other& 79.5 & 58.4 \\
 SiamTDR \cite{siamtdr} &Other& 77.2 & 55.1 \\
 LRMW \cite{lrmw} & Other & 82.5 & \textcolor{red}{\textbf{61.6}} \\
 MIRNet \cite{mirnet} & Other & 81.6 & 58.9 \\
 MBAFNet \cite{mbaf} & Other & 80.1 & 58.5 \\
 CIRNet \cite{cirnet} & Other & 81.0 & 54.4 \\
ADRNet \cite{adrnet}  & Other & 80.7 & 57.0 \\
 MANet++ \cite{manet++} & Other & 79.5 & 55.9 \\
 M5L \cite{m5l}  & Other & 79.5 & 54.2 \\
 JMMAC \cite{jmmac} & Other & 79.0 & 57.4 \\
\bottomrule
\end{tabular}

\end{table}

On RGBT234, we compare our proposed method with other 24 state-of-the-art methods. Similar to evaluation on LasHer, we also divide all 26 methods into three categories, including 5 latest prompt fine-tuning methods, 9 cutting-edge full fine-tuning methods and 12 non-fine-tuning methods.

The evaluation and comparison results are shown in Table \ref{rgbt234_result}. (1) Compared to similar methods, our M3PT seperately ranks first and second on MPR and MSR metrics, while surpasses our baseline ViPT-deep by \textbf{2.9} and \textbf{2.1} percentage points on MPR, respectively. (2) In comparison with other types of methods, our prompt method also perform great competitiveness: on the MPR metric, M3PT notably outperforms the third-ranked full fine-tuning method and the second-ranked non-fine-tuning method, while on the MSR metric, it exceeds the performance of the second-ranked full fine-tuning method and all non-fine-tuning methods. Also, We use RGBT234 toolkit to plot the MSR and MPR evaluation curves of our method and other 6 open-source methods, as shown in Figure \ref{rgbt234}.

RGBT234 provides 12 annotated attributes, including a general attribute-NO-and 11 challenging attributes: CM, HO, FM, SV, PO, LI, TC, BC, MB, DEF, LR. We operate attribute-wise MPR evaluation, as shown in Figure \ref{rgbt234_attr}. The conclusion could be summarized below: (1) In the general attribute, most trackers could precisely track the target and achieve a good score without challenges, therefore, our M3PT does not exhibit an obviously leading performance. (2) Our method shows leading performance on all 11 annotated challenging attributes (FM, MB, BC, TC, PO, SV, HO, DEF, LI, CM), demonstrating the generalization and application potential of our prompt-based tracking approach in open world scenarios. (3) We also observe that our method exhibits divergent performance on the same attributes across RGBT234 and LasHeR. For instance, concerning the FM and BC attributes in RGBT234, our method’s improvement relative to the baseline is not as pronounced as observed in LasHeR. This discrepancy may be ascribed to divergent training sets employed for model evaluation across the two benchmarks. Additionally, observing that our baseline and other sota methods exhibit similar phenomenon, we think variations in data distribution between the two benchmarks, such as the presence of different object categories and the cross-distribution of distinct attributes within the same sequences, might also contribute to the inconsistency.

\subsection{Evaluation Results on RGBT210}

\begin{table}[h]
\caption{The overall performance on RGBT210. The best two results among each type of methods are shown in \textcolor{red}{\textbf{red}} and \textcolor{blue}{\textbf{blue}} fonts, respectively.}\label{rgbt210_result}
\begin{tabular}{@{}llll@{}}
\toprule
 Method & Type & PR$\uparrow $ & SR$\uparrow $ \\
 \midrule
M3PT (Ours) &Prompt Fine-tune & \textcolor{red}{\textbf{85.0}} & \textcolor{red}{\textbf{61.3}}  \\
UnTrack \cite{untrack} &Prompt Fine-tune & 81.5 & 59.0 \\
ViPT-deep \cite{vipt} &Prompt Fine-tune & 81.3 & 59.2 \\
ProTrack \cite{protrack} &Prompt Fine-tune & 69.7 & 52.0 \\
\midrule
TBSI \cite{tbsi} &Full Fine-tune & \textcolor{red}{\textbf{85.3}} & \textcolor{red}{\textbf{62.5}} \\
DMCNet \cite{dmcnet} &Full Fine-tune & 79.7 & 55.5 \\
HMFT \cite{vtuav} &Full Fine-tune & 77.7 & 55.7 \\
\midrule
SiamMLAA \cite{siammlaa} &Other & 77.9 & \textcolor{red}{\textbf{56.7}} \\
CIRNet \cite{cirnet} &Other & 78.9 & 52.3 \\
ADRNet \cite{adrnet} &Other & \textcolor{red}{\textbf{79.4}} & 54.8 \\
JMMAC \cite{jmmac} &Other & 78.3 & 55.3 \\
MANet++ \cite{manet++} &Other & 77.2 & 53.4 \\
CAT \cite{cat} &Other & 79.2 & 53.3 \\
\bottomrule
\end{tabular}
\end{table}

On RGBT210, We compare our method with other 12 state-of-the-art methods, including 3 prompt fine-tuning methods, 3 full fine-tuning methods and 6 non-fine-tuning methods. As shown in Table \ref{rgbt210_result}, our method surpasses the baseline ViPT-deep by \textbf{3.7} and \textbf{2.1} percentage points, and ranks first among prompt methods on all metrics. Furthermore, it demonstrates a PR score that closely approaches the best full fine-tuning method. The SR and PR evaluation curves are shown in Figure \ref{rgbt210}.

\subsection{Evaluation Results on GTOT}

\begin{table}[h]
\caption{The overall performance on GTOT. The best result among each type of methods are shown in \textcolor{red}{\textbf{red}} fonts.}\label{gtot_result}
\begin{tabular}{@{}llll@{}}
\toprule
 Method & Type & PR$\uparrow $ & SR$\uparrow $ \\
 \midrule
M3PT (Ours) &Prompt Fine-tune & \textcolor{red}{\textbf{92.7}} & \textcolor{red}{\textbf{76.9}}  \\
ViPT-deep \cite{vipt} &Prompt Fine-tune & 91.8 & 76.1 \\
ProTrack \cite{protrack} &Prompt Fine-tune & 80.8 & 68.9 \\
\midrule
USTrack \cite{ustrack} &Full Fine-tune & \textcolor{red}{\textbf{93.4}} & 78.3 \\
SiamIVFN \cite{siamivfn} &Full Fine-tune & 91.5 & \textcolor{red}{\textbf{79.3}} \\
ECMD \cite{cmd} &Full Fine-tune & 90.7 & 73.5 \\
DRGCNet \cite{drgcnet} &Full Fine-tune & 90.5 & 73.5 \\
DMCNet \cite{dmcnet} &Full Fine-tune & 90.9 & 73.3 \\
HMFT \cite{vtuav} &Full Fine-tune & 91.2 & 74.9 \\
APFNet \cite{apfnet} &Full Fine-tune & 90.5 & 73.7 \\
\midrule
 SiamMLAA \cite{siammlaa} &Other& \textcolor{red}{\textbf{91.3}} & 75.1 \\
 SiamTDR \cite{siamtdr} &Other& 88.5 & 71.4 \\
SiamAFTS \cite{siamafts} &Other& 84.9 & \textcolor{red}{\textbf{77.7}} \\
MIRNet \cite{mirnet} &Other& 90.9 & 74.4 \\
CIRNet \cite{cirnet} &Other& 90.1 & 72.8 \\
ADRNet \cite{adrnet} &Other & 90.4 & 73.9 \\
JMMAC \cite{jmmac} &Other & 90.2 & 73.2 \\
MANet++ \cite{manet++} &Other & 88.2 & 70.7 \\
CAT \cite{cat} &Other & 88.9 & 71.7 \\
\bottomrule
\end{tabular}
\end{table}

On GTOT, We compare our method with other 17 state-of-the-art methods, including 2 prompt fine-tuning methods, 6 full fine-tuning methods and 9 other methods. Our M3PT reaches 92.7 PR percentage points and 76.9 SR percentage points. When reaching the best result among prompt fine-tuning methods, M3PT also outperforms the best non-fine-tuning method and the second-best full fine-tuning in terms of PR, as shown in Table \ref{gtot_result}. We also draw the evaluating metric plots of 10 open-source methods, as shown in Figure \ref{gtot}.

\subsection{Evaluation Results on VTUAV}

\begin{table*}[h]
\caption{The overall performance on VTUAV. The best result among each type of methods are shown in \textcolor{red}{\textbf{red}} fonts.}\label{vtuav_result}
\begin{tabular}{@{}llllll@{}}
\toprule
&Type&\multicolumn{2}{c}{Long-term}&\multicolumn{2}{c}{Short-term}\\
 & & MPR$\uparrow $ & MSR$\uparrow $ & MPR$\uparrow $ & MSR$\uparrow $ \\
 \midrule
M3PT (Ours) &Prompt Fine-tune & \textcolor{red}{\textbf{55.6}} & \textcolor{red}{\textbf{47.4}} & \textcolor{red}{\textbf{75.5}} & \textcolor{red}{\textbf{63.0}}  \\
ViPT-deep \cite{vipt} & Prompt Fine-tune & 54.1 & 46.5 & 74.6 & 62.3 \\
ProTrack \cite{protrack} & Prompt Fine-tune & 46.7 & 41.7 & 72.5 & 62.9 \\
 \midrule
HMFT-LT \cite{vtuav} &Full Fine-tune & \textcolor{red}{\textbf{53.6}} & \textcolor{red}{\textbf{46.1}} & - & - \\
HMFT \cite{vtuav} &Full Fine-tune & 41.4 & 35.5 & \textcolor{red}{\textbf{75.8}} & \textcolor{red}{\textbf{62.7}} \\
mfDiMP \cite{mfdimp} &Full Fine-tune & 31.5 & 27.2 & 67.3 & 55.4 \\
\midrule
FSRPN \cite{fsrpn} &Other & \textcolor{red}{\textbf{36.6}} & \textcolor{red}{\textbf{31.4}} & \textcolor{red}{\textbf{65.3}} & \textcolor{red}{\textbf{54.4}} \\
DAFNet \cite{dafnet} &Other & 25.3 & 18.8 & 62.0 & 45.8 \\
ADRNet \cite{adrnet} &Other & 23.5 & 17.5 & 62.2 & 46.6 \\
\bottomrule
\end{tabular}

\end{table*}

VTUAV is the first benchmark to propose a long-term tracking subtask. To evaluate our method on this benchmark, we select 8 methods, including 2 prompt fine-tuning methods, 3 full fine-tuning methods and 3 none-fine-tuning methods, and compare them with our M3PT on long-term and short-term subsets, as shown in Table \ref{vtuav_result}. In addition, we plot MPR and MSR evaluation curves on two subsets by official tookit, respectively, as shown in Figure \ref{vtuavlt} and Figure \ref{vtuavst}.

It is encouraging that, despite not introducing any model online update or template update operation in the inference stage, our method still achieves state-of-the-art among all the methods on long-term subset, while the best full fine-tuning method, HMFT-LT, introduces an online parameter-update mechanism to adapt to the long-term tracking task. This is because our proposed method could fully explore the complementary information of the dual modality and maximizes the use of excellent upstream knowledge, ensuring robust performance when coping with long-term challenge.

\subsection{Evaluation Results on VOT2019RGB-TIR}

\begin{table}[h]
\caption{The overall performance on VOT2019RGB-TIR. The best result among each type of methods are shown in \textcolor{red}{\textbf{red}} fonts.}\label{vot2019_result}
\begin{tabular}{@{}lllll@{}}
\toprule
 Method & Type & A$\uparrow $ & R$\uparrow $ & EAO$\uparrow $ \\
 \midrule
M3PT (Ours) &Prompt Fine-tune & \textcolor{red}{\textbf{67.2}} & \textcolor{red}{\textbf{81.3}} & \textcolor{red}{\textbf{41.6}}  \\
ViPT-deep \cite{vipt} &Prompt Fine-tune & 66.1 & 73.1 & 34.8 \\
ProTrack \cite{protrack} &Prompt Fine-tune & 63.7 & 71.0 & 34.2 \\
\midrule
DMCNet \cite{dmcnet} &Full Fine-tune & 60.1 & 70.9 &38.0 \\
SiamCDA \cite{siamcda} &Full Fine-tune & \textcolor{red}{\textbf{68.2}} & \textcolor{red}{\textbf{75.7}} & \textcolor{red}{\textbf{42.4}} \\
\midrule
JMMAC \cite{jmmac} &Other & \textcolor{red}{\textbf{66.0}} & \textcolor{red}{\textbf{82.4}} & \textcolor{red}{\textbf{49.8}} \\
TFNet \cite{tfnet} &Other & 46.2 & 59.4 & 28.8\\
ADRNet \cite{adrnet} &Other & 62.2 & 75.7 & 39.6 \\
MANet++ \cite{manet++} &Other & 50.9 & 53.8 & 27.2 \\
FSRPN \cite{fsrpn} &Other & 63.6 & 70.7 & 35.5\\
\bottomrule
\end{tabular}
\end{table}

Different with other mainstream benchmarks, VOT2019RGB-TIR introduce EAO, Robustness and Accuracy metrics to evaluate the performance of RGB-T Trackers. On VOT2019RGB-TIR, we compare our method with other 9 state-of-the-art methods, including 2 prompt fine-tuning methods, 2 full fine-tuning methods and 5 other methods.

As illustrated in Table \ref{vot2019_result}, our M3PT achieve 67.2, 81.3 and 41.6 percentage points on Accuracy, Robustness and EAO metrics, respectively. On Robustness and EAO, Our approach surpasses the baseline ViPT-deep by \textbf{8.2}$\%$ and \textbf{6.8}$\%$ respectively, which verifies the tracking stability and reliability of our method. Additional, M3PT completely beats the first-place full fine-tuning method SiamCDA on Robustness. And our method also beats the best non-fine-tuning method JMMAC on Accuracy, although JMMAC introduces extra inter-frame motion prediction.

\subsection{Fine-tuned Parameters Comparison}

\begin{table*}[h]
\caption{The parameters comparison of different fine-tuning methods. Params represents the number of whole parameters of the model, and Tuned Params represents the number of unfrozen parameters involved in fine-tuning.}
\label{params}
\begin{tabular}{@{}llllll@{}}
\toprule
\multirow{2}*{Method}& \multirow{2}*{Type}& \multirow{2}*{Total Params}&\multirow{2}*{Tuned Params}&\multicolumn{2}{c}{LasHer}\\
 &  & & & PR$\uparrow $ & SR$\uparrow $ \\
 \midrule
 RGB Tracker & Foundation model & 92.12M & - & 51.5 & 41.2\\
\midrule
ViPT-shaw \cite{vipt} &Prompt Fine-tune & 92.73M & 0.61M (0.66$\%$) & 59.0 & 47.6\\
ViPT-deep \cite{vipt} &Prompt Fine-tune & 92.96M & 0.84M (0.91$\%$) & 64.8 & 52.2\\
UnTrack \cite{vipt} &Prompt Fine-tune & 98.72M & 6.60M (7.16$\%$) & 66.7 & 53.6\\
OneTrack \cite{vipt} &Prompt Fine-tune & 94.92M & 2.80M (3.04$\%$) & 67.1 & 54.1\\
M3PT (Ours) &Prompt Fine-tune & 92.46M & \textbf{0.34M (0.37$\%$)} & \textbf{70.0} & \textbf{56.1}\\
\midrule
TBSI \cite{tbsi} &Full Fine-tune & 201.98M & 201.98M (219.26$\%$) & 69.2 & 55.6\\
USTrack \cite{ustrack} &Full Fine-tune & 97.36M & 97.36M (101.13$\%$) & - & -\\
\bottomrule
\end{tabular}
\end{table*}

To verify the parameter-efficiency of our prompt fine-tuning method, we compare it with other recent prompt tine-tuning methods and state-of-the-art full-parameter fine-tuning methods again in terms of parameters and performance. To ensure a fair comparison, all fine-tuning method are built upon the same RGB tracking method \cite{ostrack}.

Compared to full fine-tuning methods, the major advantage of prompt fine-tuning is the great parameter-efficiency, which means the upstream model could sufficiently inherit upstream knowledge with only a small amount of tuned parameters. As shown in Table \ref{params}, Each prompt fine-tuning method adjusts less than 10$\%$ of the foundation model’s parameters, a testament to parameter-efficiency that significantly reduces the computational and storage burdens on training equipment compared to fine-tuning methods. However, for the huge gap in tuned parameters, it's usually hard for prompt fine-tuning methods to compete with sota full fine-tuning methods in performance. Nevertheless, our M3PT, surpassing all the existing prompt methods, beats the state-of-the-art full fine-tuning method on LasHer. It's remarkable that our method only has 0.34M tuned parameters, which are 0.37$\%$ of the foundation model’s parameters. The comparison results prove the superiority of our method.

\subsection{Ablation Study}

\begin{table}[h]
\caption{The ablation experiment results. In the table, the four abbreviations represent the four prompting strategies. UIEPS: Uni-modal and Inter-modal Exploration Prompt Strategy, MFPS: Middle Fusion Prompt Strategy, FEPS: Fusion-modal Enhancing Prompt Strategy, MSPS: Modality-aware and Stage-aware Prompt Strategy.}
\label{abla}
\begin{tabular}{@{}lllllll@{}}
\toprule
\multicolumn{2}{c}{UIEPS}&\multirow{2}*{MFPS}&\multirow{2}*{FEPS} & \multirow{2}*{MSPS}&\multicolumn{2}{c}{RGBT234}\\
UEP & IP &  &  &  & MPR$\uparrow $ & MSR$\uparrow $ \\
 \midrule
&  &  &  &  & 69.5 & 52.1 \\
$\surd$ &  &    &  &  & 74.9&56.5\\
& $\surd$  &    &  &  & 80.2&59.8\\
 $\surd$ & $\surd$  &    &  &  & 83.4&61.7\\
& & $\surd$  &  &  & 76.7&56.8\\
& &  & $\surd$  &  & 72.3&54.2\\
 $\surd$ & $\surd$ & $\surd$ & $\surd$  &  & 85.1 & 63.0\\
 $\surd$ & $\surd$ & $\surd$  & $\surd$  & $\surd$  & \textbf{86.5} & \textbf{63.9}\\
\bottomrule
\end{tabular}
\end{table}

To verify the effectiveness of our proposed four prompt strategies, we conduct extensive ablation studies and analyses on RGBT234. In the experiments, we build a baseline tracker which is only composed of upstream parameters, with MFP replaced by element-wise addition operation of dual-modal features. Based on the baseline, we explore the effectiveness of different prompt strategies. The ablation results are shown in Table \ref{abla}.

\textbf{Uni-modal and Inter-modal Exploration Prompt Strategy:} We investigate the impact of UEP, IP and complete Uni-modal and Inter-modal Exploration Prompt Strategy on our M3PT. As shown in Table \ref{abla}, UEP brings a guidance for the upstream backbone to extract comprehensive uni-modal features while IP leads to effective inter-modal communication by channel-level prompts. Compared with base tracker, the collaboration of UEP and IP improves the MPR and MSR of base tracker on RGBT234 by \textbf{13.9} and \textbf{9.6} percentage points, respectively, which proves that this strategy significantly guides the backbone to fully explore modality-shared features and modality-independent features.

\textbf{Middle Fusion Prompt Strategy:} We only utilize MFP on the base tracker, to verify the effectiveness of this strategy. According to Table \ref{abla}, the model only with Middle Fusion Prompt Strategy surpasses base tracker by \textbf{7.2} and \textbf{4.7} percentage points on the MPR and MSR metrics on RGBT234, respectively. This indicates that this prompt strategy can achieve adaptive selection of discriminative features, providing reliable fusion prompts for the second-stage backbone. 

\textbf{Fusion-modal Enhancing Prompt Strategy:} We validate Fusion-modal Enhancing Prompting Strategy. Table \ref{abla} shows that this strategy boosts the MPR and MSR metrics of the original tracker on RGBT234 by \textbf{2.8} and \textbf{2.1} percentage points, respectively. This confirms that the strategy better adapts the second-stage backbone to fusion-modal modeling and feature enhancement.

\textbf{Modality-aware and Stage-aware Prompt Strategy:} To evaluate the effectiveness and necessity of the Modality-aware and Stage-aware Prompt strategy, we build a tracker that contains the other three prompt strategies and a tracker that contains all four prompt strategies. As shown in Table \ref{abla}, although the collaboration of the three prompt strategies for different purposes has already achieved excellent performance on RGBT234, this strategy still improves MPR and MSR metrics by \textbf{1.4} and \textbf{0.9} points, respectively. With only a small amount of parameters, the strategy stores low-level fixed patterns within the modality and enables the foundation model to better identify current modality and sub-task.

\subsection{Variable Analysis}

\begin{table}[h]
\caption{Fusion location in our proposed middle fusion tracking framework.}\label{middle fusion}
\begin{tabular}{@{}llll@{}}
\toprule
Fusion Location &MPR$\uparrow $&MSR$\uparrow $&Speed$\uparrow $\\
 \midrule
1  &62.1 & 47.3 & \textbf{94.7fps} \\
6  & 67.8 & 51.4 & 71.3fps \\
11  & \textbf{73.5} & \textbf{55.2} & 59.1fps \\
\bottomrule
\end{tabular}
\end{table}

\textbf{Variable in Middle Fusion Framework.} We explore the impact of different fusion location value on speed and performance of tracker with our middle fusion meta-framework. First, we remove all prompt modules like the base tracker in ablation study, to prevent the influence of other components. From Table \ref{middle fusion}, the tracker operates at a fast speed with marginally inferior performance when the fusion module is inserted after the 1st layer. Conversely, the inference speed yields to enhanced performance when the fusion module is inserted after the 11th layer. When the fusion location is set to 6, the tracker’s performance and speed have achieved a harmonious balance. These results show our Middle fusion Framework's flexibility, indicating that we could adjust the variable fusion location to strike an relatively ideal balance between effectiveness and efficiency, thereby meeting the diverse requirements of various applying scenarios.

\begin{table}[h]
\caption{First-stage block number in our proposed M3PT.}\label{block nums}
\begin{tabular}{@{}llll@{}}
\toprule
Block Number & MPR$\uparrow $ & MSR$\uparrow $ & Speed$\uparrow $\\
 \midrule
3  & 85.1 & 62.9 & \textbf{44.6fps} \\
6  & 85.5 & 63.3 & 39.9fps \\
10  & \textbf{86.5} & \textbf{63.9} & 34.1fps \\
\bottomrule
\end{tabular}
\end{table}

\textbf{First-Stage Block Number of M3PT.} Considering our method sets an example for application of the proposed middle fusion meta-framework, we also analysis the impact of the first-stage block number on our M3PT. From Table \ref{block nums}, adding the number of the first-stage blocks augments the extraction of uni-modal information and fosters a more effective complementary interplay between the two modalities, thereby refining the foundational model’s adaptation to the downstream task, albeit at the expense of efficiency. The findings verify the flexibility of our fusion framework again.

\subsection{Qualitative Analysis}

From LasHer, We conduct a qualitative analysis of our proposed method on three video sequences, which contain four common challenging scenarios: low illumination and illumination variation, which degrade the reliability of visible modality; similar appearance, which is a common challenge for tracking tasks; and temperature crossover, which degrades the reliability of thermal infrared modality. We compare our method with four state-of-the-art RGB-T tracking methods: ViPT-deep \cite{vipt}, ProTrack \cite{protrack}, APFNet \cite{apfnet} and HMFT \cite{vtuav}.

According to the qualitative comparison results in Figure \ref{all} (a)-(d), our method’s tracking results are the most stable and accurate in low illumination and illumination variation scenarios, fully exploiting the potential of RGB-T prompt tracking. In the similar appearance scenario, our method can still accurately locate and estimate the target boundary, while other methods fail due to interference from similar objects. In the temperature crossover scenario, which is a unique challenge for RGB-T tracking, our method is clearly the closest to the ground truth. The qualitative test results demonstrate the high robustness of M3PT in various common challenging scenarios.

\begin{figure}
    \centering
    \includegraphics[width=1\linewidth]{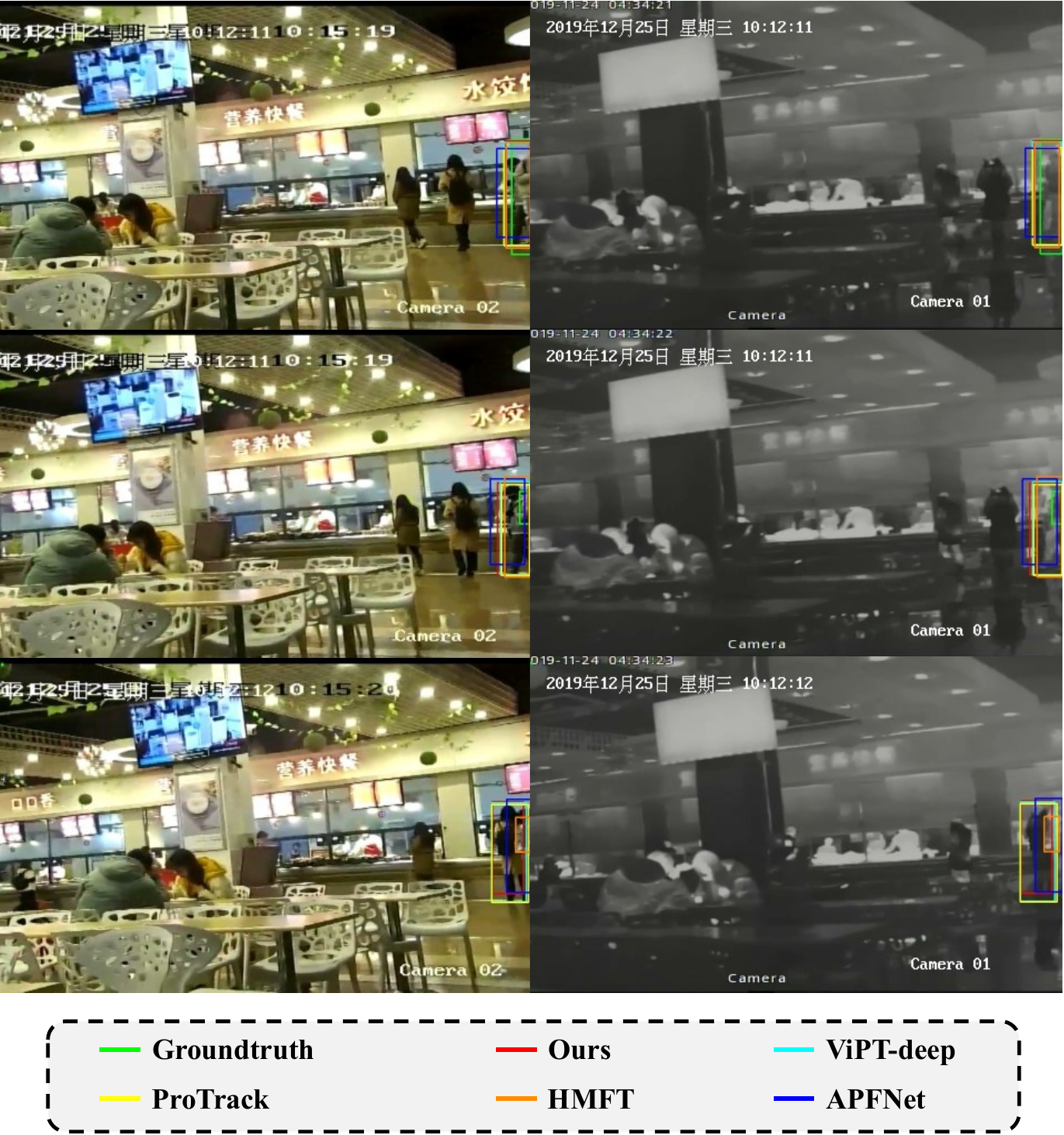}
    \caption{A failure case in an out-of-view scenario.}
    \label{fail}
\end{figure}

Furthermore, corresponding to our attribute-wise evaluation results, we present a tracking failure case, containing an out-of-view scenario, to better analyse the reason of failure. As depicted in Figure \ref{fail}, when the target disappears from view, the tracker is constrained to re-localize, visually similar interfering objects across two modalities. This ultimately culminates in the accumulation of model errors and the subsequent failure of tracking.

\section{Conclusion}

In this work, we propose a novel prompt tracking method based on a middle fusion meta-framework and using four flexible visual prompt strategies. The four prompt strategies not only maximize the use of upstream knowledge, but also fully exploit the potential of prompt learning in RGB-T tracking tasks. Evaluation results on six recent challenging RGB-T tracking benchmarks demonstrate the superiority of our method over the current best prompt fine-tuning tracking methods, while also showcasing competitive performance compared to most full fine-tuning methods. Moreover, we offer new insights for the exploration of visual prompt learning. In the future, we plan to involve introducing motion cues or linguistic prompts to better address extreme scenarios. Furthermore, we intend to extend our parameter-efficient prompting approach to various other multi-modal video understanding tasks.

\section*{Acknowledgements}

This work was supported in part by the National Natural Science Foundation of China under Grant 62273048.










\bibliographystyle{unsrt}

\bibliography{cas-refs}



\end{document}